# Quantum Model Parallelism for MRI-Based Classification of Alzheimer's Disease Stages

Emine Akpinar[1*], Murat Oduncuoglu[1]

[1]Department of Physics, Yildiz Technical University, Istanbul, Turkey

*Corresponding author: Emine Akpinar, emineakpinar28@gmail.com

## Abstract

With increasing life expectancy, Alzheimer's disease (AD) has become a major global health concern. While classical AI-based methods have been developed for early diagnosis and stage classification of AD, growing data volumes and limited computational resources necessitate faster, more efficient approaches. Quantum-based AI methods, which leverage superposition and entanglement principles along with high-dimensional Hilbert space, can surpass classical approaches' limitations and offer higher accuracy for high-dimensional, heterogeneous, and noisy data. In this study, a Quantum-Based Parallel Model (QBPM) architecture is proposed for the efficient classification of AD stages using MRI datasets, inspired by the principles of classical model parallelism. The proposed model leverages quantum advantages by employing two distinct quantum circuits, each incorporating rotational and entanglement blocks, running in parallel on the same quantum simulator. The classification performance of the model was evaluated on two different datasets to assess its overall robustness and generalization capability. Additionally, gradient variance analysis empirically and theoretically confirmed that the architecture remains free from barren plateaus, ensuring model trainability as qubit counts scale. Results obtained under high Gaussian noise provided experimental evidence for the model's applicability not only in theoretical but also in practical scenarios. Moreover, compared with five classical transfer learning methods, the proposed model demonstrated its efficiency as an alternative to classical approaches by achieving higher classification accuracy and comparable execution time while utilizing significantly fewer circuit parameters. The results indicate that the proposed QBPM architecture represents an innovative and powerful approach for the classification of stages in complex diseases such as Alzheimer's.

### 1. Introduction

The brain is highly vulnerable to neurological diseases and infections affecting neurons or tissues, and any structural or functional damage can impact the whole body. In this context, neurodegenerative diseases in the central nervous system, which are often accompanied by irreversible cell loss, have emerged as one of the most critical challenges in modern medicine, particularly due to their increasing prevalence with the aging population [1–3]. Alzheimer's disease (AD), a progressive brain disorder marked by the gradual degeneration or death of neurons, affects millions of individuals worldwide and leads to cognitive and behavioral impairments that profoundly interfere with daily functioning [4–7]. Alzheimer's Disease International (ADI) reports that while more than 55 million individuals had dementia worldwide in 2020, this figure is projected to increase to 78 million by 2030 and 139 million by 2050 [8]. AD primarily arises from the irreversible degeneration of nerve cells, the accumulation of amyloid-beta (Aβ) peptides in amyloid plaques within the brain parenchyma, and the aggregation of tau protein in the form of neurofibrillary tangles in neurons [9,10]. The stages of the disease include the preclinical (asymptomatic) stage, mild cognitive impairment, and dementia, with dementia itself being subdivided into three stages: mild, moderate, and severe. In the initial phases, mild cognitive impairments are observed, whereas as the disease progresses, severe symptoms such as anxiety, depression, and insomnia may develop; in more advanced stages, even basic tasks such as walking and eating become significantly impaired. Furthermore, since AD-related cognitive impairments are often perceived as a normal aspect of aging, the disease typically remains undetected until its later stages [11,12]. However, if individuals with mild cognitive impairment can be identified at an early stage, the progression toward AD may be delayed or even partially prevented [5,13]. While no established cure exists for AD, early detection remains vital for managing cognitive decline, alleviating symptoms, and supporting patients' daily functioning.

At present, the diagnosis of AD and the determination of its stages rely on potential biomarkers and imaging techniques. Among these biomarkers are alterations in levels associated with the accumulation of amyloid-β peptides in the brain and increases in tau protein [14]. However, these factors are not definitive or exclusive indicators. Magnetic resonance imaging (MRI) stands out as a key neuroimaging tool frequently utilized for the early detection and classification of AD. Its widespread use is attributed to its non-reliance on ionizing radiation, exceptional soft tissue contrast, and its capability to clearly depict structural alterations such as brain atrophy, gray matter loss, and the enlargement of ventricles [15,16]. Although MRI can indicate specific findings related to AD, the increasing use of this imaging technique in the aging population has led to a substantial growth in data volume, making the manual processing, analysis, and extraction of meaningful insights increasingly challenging and time-consuming [17,18]. As the amount of data grows, issues such as noise, complexity, high correlation, and dimensionality also increase, creating additional challenges in data analysis and classification processes.

Currently, computer vision and artificial intelligence (AI) technologies are frequently employed to overcome these data-driven challenges and to enable the early detection of AD from MRI scans. In other words, the early diagnosis of AD from MR images stands out as an important problem waiting to be solved in the field of computer vision. In this context, classical deep learning (DL) models based on artificial neural networks (ANNs) and convolutional neural networks (CNNs) are widely used for the extraction of discriminative features from large patient datasets and for disease classification processes based on these features, aiming to identify brain regions affected by AD at an early stage, achieving varying levels of success [19–24]. ANNs are structures composed of interconnected neurons that mimic the learning processes performed by neurons in the human nervous system, where learning occurs through information transfer between these neurons. This architecture is widely used in complex data analysis and classification problems, particularly because it can model nonlinear systems with intricate relationships between features and establish connections between variables through mathematical functions. DL models built on ANN architectures, including Recurrent Neural Networks (RNNs), Deep Neural Networks (DNNs), Multi-Layer

Perceptrons (MLPs), and Long Short-Term Memory networks (LSTMs), are primarily used to identify and categorize AD at early stages [25–27], but also for predicting disease progression [28,29] and analyzing biomarkers [30,31]. Another DL approach, CNNs, is inspired by the neocortex—a part of the human brain cortex responsible for transmitting and processing sensory signals through a complex hierarchical structure over time—and provides a layered and hierarchical architecture in which features are automatically extracted from input data. Fundamentally, a CNN model consists of input, convolutional, activation, pooling, and fully connected layers, with the process of learning features from data typically carried out by small filters in each layer, which are designed to detect specific features. In CNN-based architectures, higher-level features are defined based on lower-level features through hierarchical learning processes, and the complex architectural design of CNNs provides a degree of robustness and flexibility against shifts and transformations [32]. The literature contains numerous examples of CNN-based models developed from scratch, as well as pre-trained transfer learning (TL) architectures such as LeNet, AlexNet, VGGNet, GoogLeNet, DenseNet, and ResNet, applied for the early detection and classification of AD [17,33–38].

However, factors such as the complex structures within the data, dense pixel-level information, and high correlations make it difficult for classical models to distinguish subtle tissue differences, posing significant challenges in the classification of AD. In addition, classical DL models require large amounts of labeled data for effective training (to mitigate the risk of overfitting) [39]. Moreover, as models aim to learn more important and discriminative features from MR images, model complexity increases concurrently, significantly escalating interactions with computational resources [40]. In particular, there is a growing need for extensive computational resources beyond the limitations of conventional processing units such as CPUs and GPUs [39,41]. Recent advances and investments in hardware technologies have led to significant improvements in computational capacity; however, these developments remain insufficient when compared to the globally increasing data volumes, limiting the ability to fully exploit and utilize the data [42]. To overcome these challenges, and to better manage complex computations such as the classification of AD, as well as to maximize information extraction from data, quantum computing and quantum computing-based AI techniques offer a promising alternative approach.

Quantum computing is an innovative computational framework that leverages the postulates of quantum mechanics and the properties of quantum particles for information processing and computation, drawing upon multiple disciplines such as physics, mathematics, and computer engineering. Moreover, quantum computing enables the solution of certain problems more efficiently compared to classical computation [43–46]. In this context, information is stored in quantum states of physical structures known as qubits (or quantum bits), and information transfer is performed through these entities. At present, qubit technologies suitable for small-scale quantum computers are being developed based on various physical platforms, such as photons, superconductors, trapped ions, and topological systems [43,47,48]. Among these platforms, gate-based quantum computers employing superconducting qubits stand out in terms of accessibility and applicability due to their existing infrastructure and scalable manufacturing capabilities. In contrast to classical bits, qubits can exist in the states $|0\rangle$, $|1\rangle$, or their linear superpositions ($\alpha|0\rangle + \beta|1\rangle$), while also being capable of exhibiting quantum entanglement. These quantum properties not only allow information to be represented in a much broader space [49] but also enhance computational capacity through quantum parallelism [49,50]. Furthermore, the exponentially large Hilbert space in which quantum states reside provides significantly greater information capacity compared to classical vector spaces, as well as robustness particularly against noisy data [51,52]. However, contemporary quantum computers fall under the category of Noisy Intermediate-Scale Quantum (NISQ) devices, characterized by limited circuit depth and a restricted number of qubits, and they remain vulnerable to external perturbations and quantum decoherence, which hinders computational accuracy and the achievement of full quantum advantage. Despite these limitations, as highlighted in numerous studies, quantum computing and quantum computing-based AI methods/algorithms have emerged as promising alternatives in fields requiring complex medical data analysis, including drug discovery [53–55], tumor detection and classification [39,56–62], genome analysis [56,63,64], DNA sequencing [65], and the classification/early diagnosis of AD [66–73]. In particular, quantum algorithms provide a significant advantage in addressing challenges associated with highly correlated, complex, heterogeneous, and unevenly distributed medical datasets. Quantum machine learning (QML) represents a quantum computing-based AI strategy that merges quantum computing with conventional machine learning methods. By harnessing quantum phenomena such as superposition, entanglement, and interference, it aims to enhance the computational strength and efficiency of standard algorithms [74–77]. In particular, analogous to the operational principles of classical classification problems, the probabilistic nature of QML algorithms and their execution within a Hilbert space—which provides an exponentially larger search space compared to classical vector spaces—form the foundation of their performance advantages. In addition to QML algorithms, hybrid classical-quantum or quantum-classical neural networks (HQNNs), which integrate the benefits of both classical and quantum computing, enable the incorporation of quantum advantages into the learning capabilities of classical neural networks, particularly for complex, high-dimensional datasets such as medical data [59,78–80]. Put differently, hybrid models integrate the high-speed and complex processing capabilities of quantum computing with the reliability and adaptability of classical computation. Furthermore, HQNNs feature iterative learning processes and are often constructed by integrating problem-specific ansatzes or hardware-efficient ansatzes [81,82] in various configurations. The flexibility of ansatz designs, achieved through different quantum gates and different configurations, facilitates the parallel execution of quantum circuits and allows the integration of the "parallel processing" principle into quantum models, substantially enhancing computational efficiency. Model parallelism is a parallelization strategy in which the layers of a model (e.g., an ANN or CNN) are distributed across multiple accelerators (e.g., GPUs), with each accelerator training its corresponding portion of the model on the same dataset (batch).

In this study, the classical model parallelism principle has been transferred to the quantum computing environment; thereby enabling different circuit structures to capture important patterns and features in MR images of AD — such as brain atrophy, gray matter loss, and ventricular enlargement — more clearly, thus allowing AD stages to be distinguished from each other with higher accuracy rates.

The main contributions of this study to the literature are as follows:

- To the best of our knowledge, this study represents the first comprehensive investigation in the literature on the application of the Quantum-Based Parallel Model (QBPM) architecture, developed by drawing inspiration from the principle of classical model parallelism, for the classification of AD stages.
- The performance of the proposed model was evaluated on both the original OASIS-1 (Open Access Series of Imaging Studies) dataset and the external validation dataset ADNI (Alzheimer's MRI Classification). Classification performance was assessed using average training/validation loss and accuracy, along with evaluation metrics such as recall, precision, and $F_1$-score. The results obtained demonstrate the overall robustness of the model and its strong learning capacity, while, its performance across various patient profiles confirms its generalization capability.

- In addition, a comprehensive barren plateau analysis was conducted to evaluate the theoretical scalability and trainability of the proposed QBPM architecture as the qubit count increases ($n = 10,12,15$). By examining the scaling behavior of the cost function gradient variance before and after training, it has been empirically and theoretically demonstrated that the proposed architecture operates completely free from barren plateau risks across the evaluated qubit range.
- Furthermore, the proposed QBPM architecture was executed in environments where both real-world conditions and gate-induced errors occurring in real quantum computing systems were simulated with high levels of Gaussian noise. The results experimentally confirm that the model is applicable not only in theory but also under practical conditions.
- Finally, the classification performance of the proposed model was rigorously benchmarked against five classical TL architectures—EfficientNetB0, InceptionV3, MobileNetV2, ResNet50, and VGG16—under the same experimental conditions. The results clearly demonstrate that the proposed model, with significantly fewer circuit parameters, provides a strong alternative to classical models in terms of both classification accuracy and execution time.

The remainder of this study is organized as follows:

**Section 2** consists of five subsections. First, detailed information is provided regarding the datasets used in this study and the stages of AD. Next, the preprocessing steps applied to the MR images are described. In the following subsection, the mathematical structure and physical principles of the QBPM architecture, which is employed for the classification of AD stages, are presented in detail. Then, the theoretical framework and mathematical formulation of the Barren Plateau analysis, used to evaluate the trainability and scalability of the proposed architecture, are outlined. Finally, the metrics used to evaluate the classification performance of the proposed model are described. **Section 3** presents the performance of the proposed QBPM architecture in distinguishing AD stages, organized under seven subsections. In the first three subsections, the results obtained when the original dataset is fed into the model are reported, whereas in the subsequent three subsections, the results obtained using the external validation dataset are presented. In the final subsection, the results of five different classical TL methods run under the same experimental conditions are presented in detail in order to compare the classification performance of the proposed model. **Section 4** provides a comprehensive analysis and discussion of the results reported in **Section 3**, including a thorough review of the classical and quantum AI literature related to AD. Finally, **Section 5** summarizes the main findings of the study, discusses its limitations, and outlines directions for future research.

## 2. Materials and Methods

In this section, the Quantum-Based Parallel Model (QBPM) architecture, inspired by the principle of classical model parallelism and designed to distinguish between the stages (classes) of AD, is described in detail. The comprehensive workflow of the model is presented in **Figure 1**. The first part of the section provides information on the datasets used in the study and the classes corresponding to the stages of AD. This is followed by a description of the preprocessing steps applied to the data. Next, the proposed QBPM architecture is elaborated, with its two main components—quantum and classical computation—discussed in detail. Finally, the theoretical framework of the Barren Plateau analysis and the metrics used to evaluate the model's classification performance are presented, respectively.

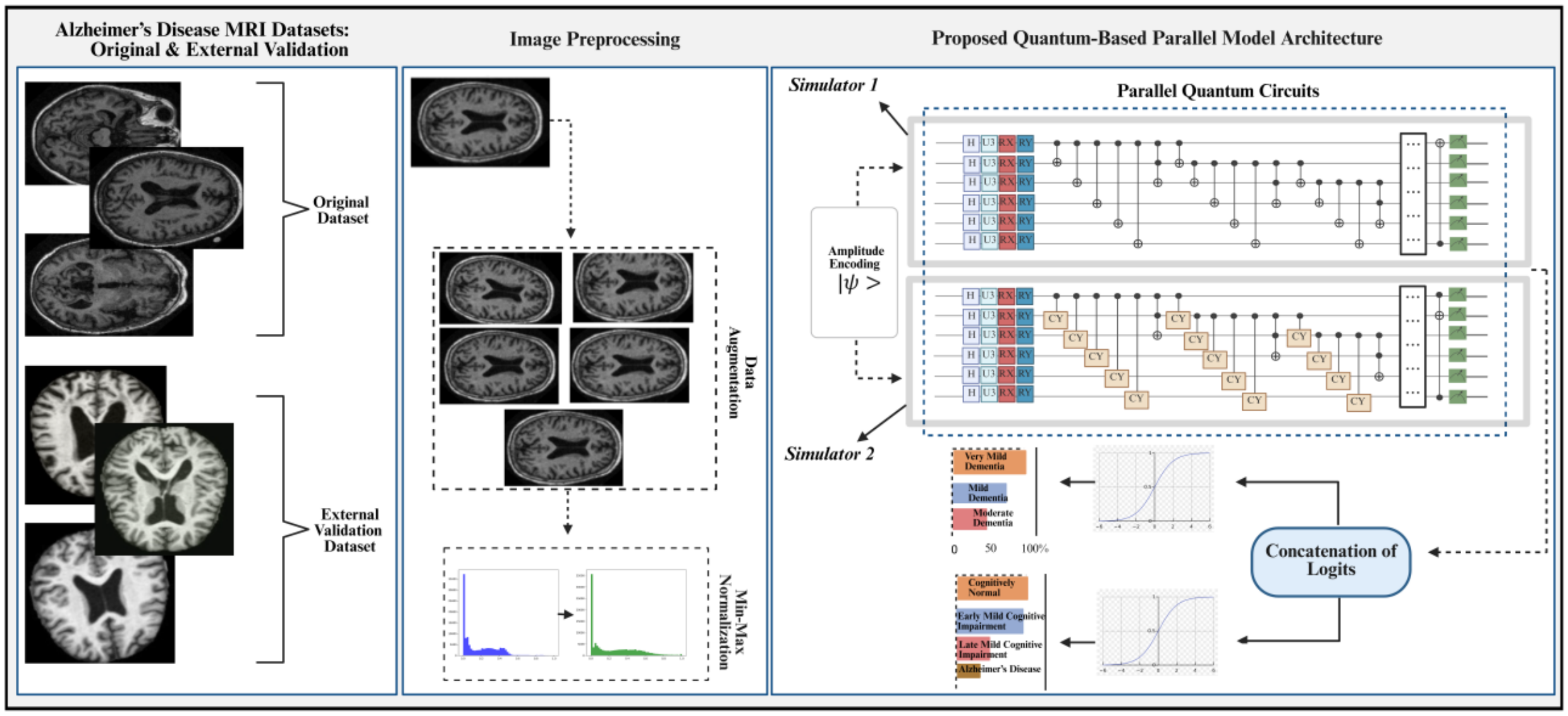

**Figure 1.** Flowchart illustrating the main steps and processing sequence of the study. In the first stage, the OASIS-1 (original) and ADNI (external validation) datasets were defined. In the second stage, class-based data augmentation and normalization preprocessing steps were performed. To provide a clearer understanding of the normalization process, the histogram distributions of a sample MR image before and after normalization are presented. In the final stage, the details of the QBPM architecture, developed to differentiate between the stages of AD, are presented. The model consists of two distinct components: quantum (amplitude encoding, Parallel PQCs, measurement) and classical (Softmax, loss computation, optimization). Here, the Parallel PQCs in fact operate simultaneously on the same quantum simulator. To emphasize the parallel structure in the flowchart, these simulators are depicted separately as "Simulator 1" and "Simulator 2."

### *2.1. Alzheimer's Disease Datasets*

The dataset consisting of brain MR images defining the very mild dementia, mild dementia, and moderate dementia stages of AD belongs to the OASIS-1 (Open Access Series of Imaging Studies) project [83] and was obtained from the Kaggle platform [84] in a processed format, having been converted from .nii to .jpg. The dataset contains a total of 86,437 MRI images, of which 67,222 belong to healthy individuals without AD, categorized under the non-demented class. The remaining images represent different stages of AD: 13,725 images correspond to very mild dementia, 5,002 to mild dementia, and 488 to moderate dementia. The details describing these stages are as follows [85,86]:

**Very Mild Dementia:** The very mild dementia stage associated with AD is a period in which the individual begins to experience difficulties in memory and thinking processes, and mild forgetfulness starts to occur. During this period, although the individual can continue to live independently, symptoms such as the recent forgetting of newly learned information, personality changes such as irritability and anger, getting lost in familiar places, and difficulties in making sound decisions affect the individual's quality of life.

**Mild Dementia:** The mild dementia stage, which is one of the longest-lasting stages in AD (which may last up to 10 years), is characterized by increasingly pronounced symptoms, and the individual requires a certain level of attention and assistance in daily living activities and personal care. During this period, personality changes are observed more frequently, and difficulties occur in mental processes such as problem-solving, planning, and decision-making.

**Moderate Dementia:** In this stage, the individual requires regular support for daily living activities and experiences forgetfulness regarding the concepts of time, person, and place. At this stage, information such as the current day or season, address, phone number, or school history may be forgotten, and family members may be confused. Additionally, personality and behavioral changes (for example, the emergence of unrealistic suspicions) can be observed during this period.

Additionally, to test the generalizability and robustness of the proposed quantum-based parallel method, the ADNI: Alzheimer's MRI Classification Dataset [87], which includes four different classes—Cognitively Normal, Early Mild Cognitive Impairment, Late Mild Cognitive Impairment, and Alzheimer's Disease—was used at this stage. In these classes, there are 6,464, 9,600, 8,960, and 8,960 sample MRI images, respectively.

The class distributions in both the original dataset and the external dataset, which was used for model validation, are presented in detail in **Figure 2a**. Four different sample MRI images for each class in these datasets are shown in **Figure 2b**.

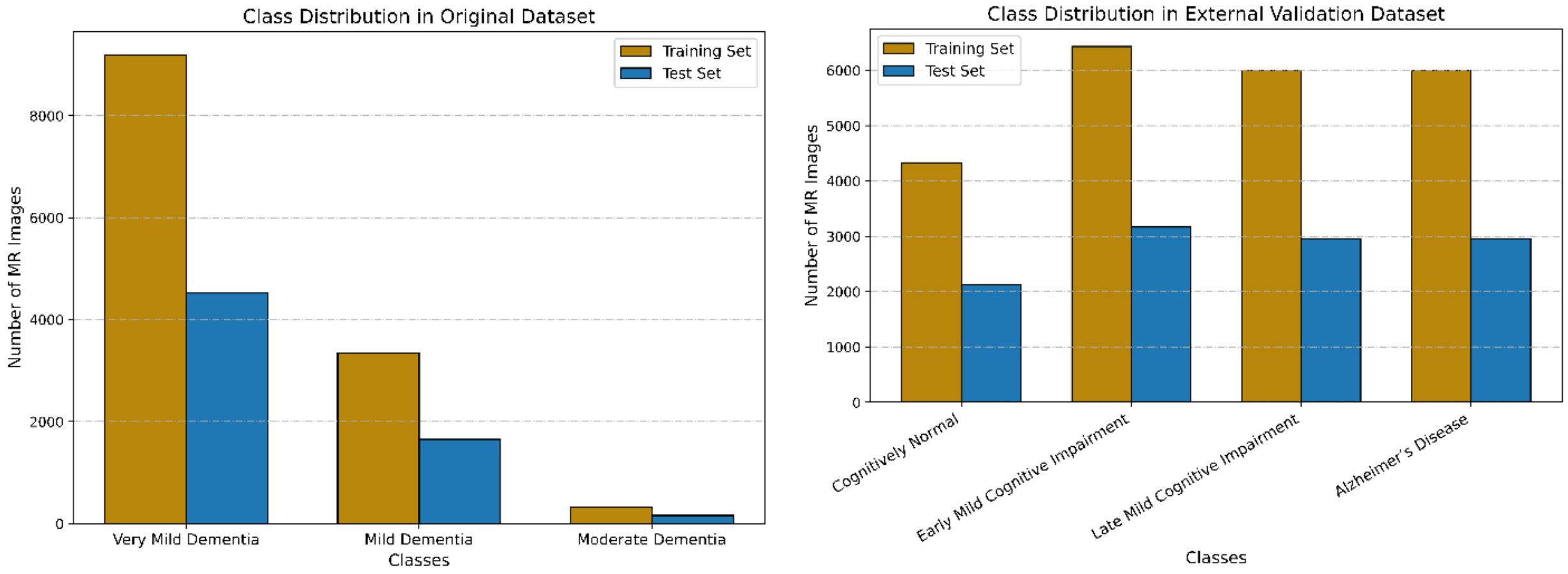

**Figure 2a.** Class distributions in the original (OASIS-1) and external validation (ADNI) datasets. In the OASIS-1 training set, there are 9,196 samples of Very Mild Dementia, 3,351 samples of Mild Dementia, and 327 samples of Moderate Dementia, while the test set contains 4,529, 1,651, and 161 samples, respectively. In the ADNI training set, the classes include 4,331 samples of Cognitively Normal, 6,432 samples of Early Mild Cognitive Impairment, 6,003 samples of Late Mild Cognitive Impairment, and 6,003 samples of Alzheimer's Disease. The test set contains 2,133, 3,168, 2,957, and 2,957 samples, respectively.

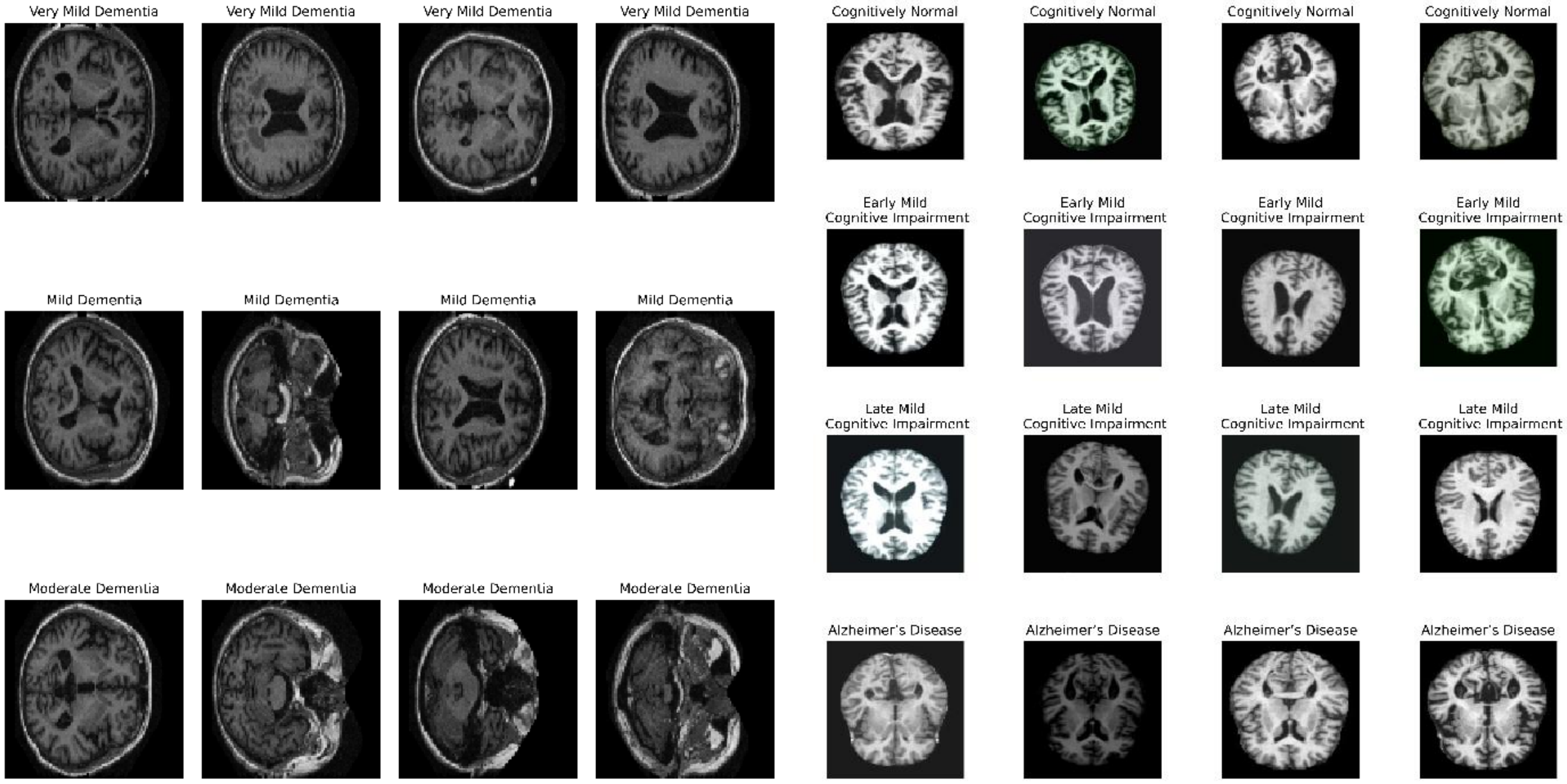

**Figure 2b.** Sample MRI images for each class in the original and external validation datasets. As an example, four different MRI images from each class are presented.

### *2.2. Image Preprocessing*

After defining the datasets, the first step involved converting the data into NumPy arrays using the OpenCV library, followed by resizing operations. During the resizing process, the dimensions of all MR images were set to (100, 100, 3). In this way, the data size was reduced minimally, information loss was kept to a minimum, and the data was prepared to be compatible with the quantum computing environment. In the next step, preprocessing was completed through normalization and data augmentation procedures. Detailed information regarding these processes is provided below.

#### *2.2.1. Normalization*

In QML studies, one of the most important steps in the preprocessing stage is the normalization process, in which the data is scaled to a specific range. This procedure ensures that both the classical and quantum components operate more stably and compatibly, while also helping to maintain the consistency of the quantum states generated in the quantum circuits. In this study, the min-max scaler method was utilized to scale each pixel value to the [0, 1] range.

#### *2.2.2. Class-Based Augmentation*

In the final step of the preprocessing stage, namely the data augmentation section, the number of MR images was increased using the ImageDataGenerator module from the Keras library. At this stage, "class-based data augmentation" procedures were performed; thus, the representation power of minority classes was enhanced, data imbalance was reduced, and the quantum model was enabled to better learn these classes. During this process, various transformations such as shifting, cropping, rotation, zooming, horizontal flipping, and filling missing regions were applied using the ImageDataGenerator module to achieve data augmentation. Here, the augmentation process was applied only to the Moderate Dementia MRI images, which constitute the minority class in the original dataset. Since the external validation dataset exhibited a relatively more balanced distribution compared to the original dataset, no augmentation was performed on this dataset. The data augmentation procedures applied to a sample MRI image in the minority class of the original dataset are shown in **Figure 3**.

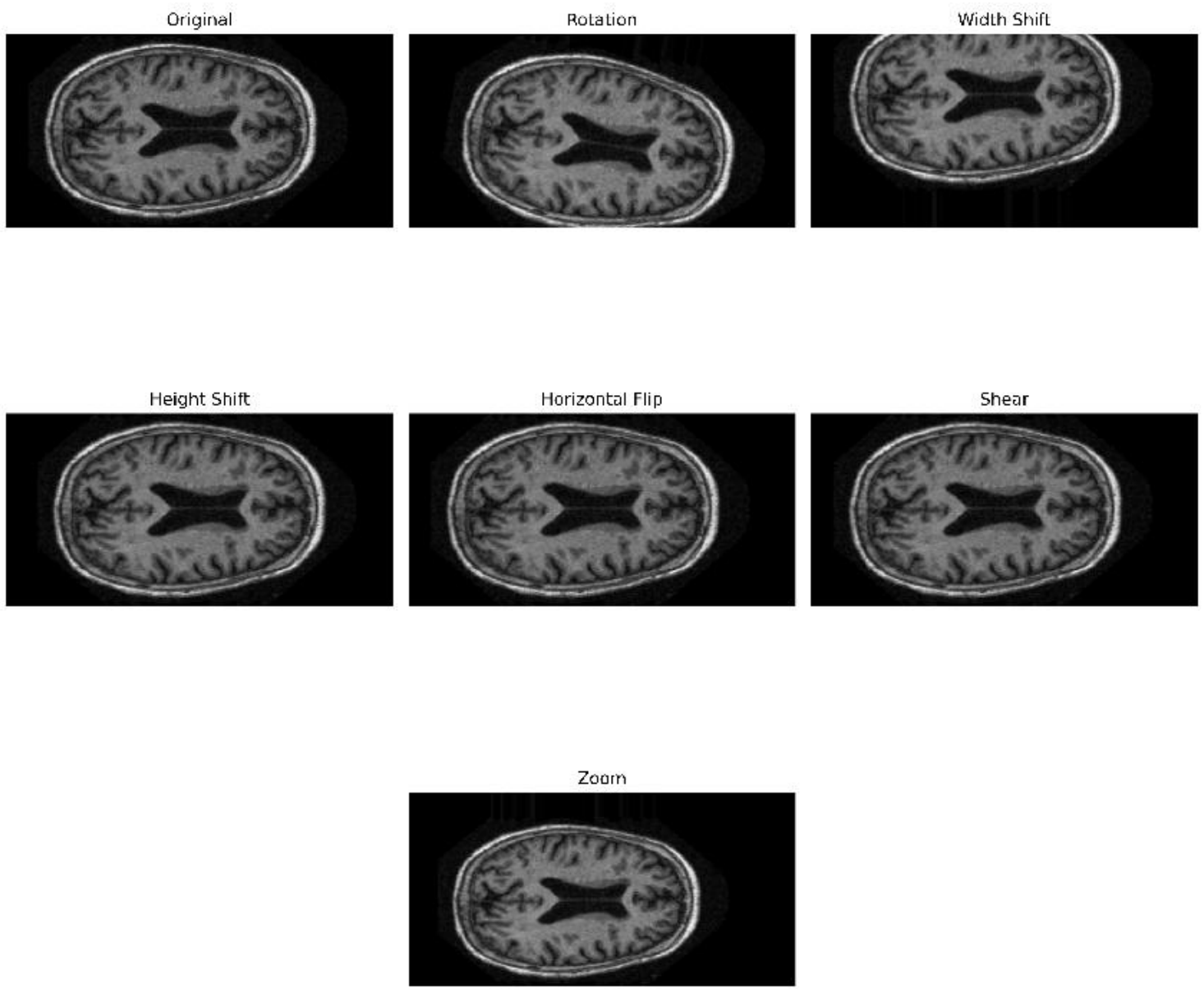

**Figure 3.** Data augmentation procedures applied to an MR image belonging to the Moderate Dementia minority class in the original dataset. In this process, the values of rotation range, width shift range, height shift range, shear range, and zoom range were set to 20, 0.2, 0.2, 0.2, and 0.2, respectively. Additionally, horizontal flipping was applied, and the fill mode was set to 'nearest'.

### *2.3. Proposed Quantum-Based Parallel Model Architecture*

In this study, the Quantum-Based Parallel Model (QBPM) architecture, developed by drawing inspiration from the classical model parallelism principle to classify the stages of AD, consists of two main components. The first component represents the quantum computing part of the study and includes feature mapping, parallel parameterized quantum circuits (Parallel PQCs), and measurement stages. The classical component, on the other hand, encompasses the softmax function, loss computation, and optimization steps.

Quantum Computing

#### *2.3.1. Feature mapping*

Since quantum algorithms and quantum computers can only interact with or operate on data expressed as quantum states, classical data must be represented in the quantum space. This enables a rich data representation and allows tasks such as learning complex patterns and class separation to be performed more effectively. Thus, quantum data, expressed as quantum bits or qubits, are capable of existing in multiple states at the same time, thereby enabling a wide range of information to be processed in parallel and efficiently [88]. Furthermore, due to the entanglement property, the state of one qubit can be correlated with the state of another qubit regardless of distance, enabling the correlation information in the data to be represented more effectively through quantum states [89,90]. This is particularly important for representing features in MRI data with complex and high correlations, such as brain atrophy, gray matter loss, and ventricular enlargement in AD, more effectively in quantum space compared to classical vector space. In the quantum feature mapping process, the data is projected from the classical vector space to the high-dimensional, nonlinear Hilbert space—i.e., the quantum feature space—where the inner product is defined, thereby achieving a richer data representation [52,91,92]. This process is expressed as $\mathbb{D} \rightarrow \mathbb{W}$, where $\mathbb{D}$ represents the dataset defined in the classical vector space. Here, $\mathbb{D} = \left(x^d,\ y^d\right)_{d=1}^{D}$, and for an $N-$dimensional feature vector, $x^d \in \mathbb{R}^N$ and $\mathbb{D} \subset \mathbb{R}^N$. $\mathbb{W}$, on the other hand, is a subset of the complex vector space ($\mathbb{C}^n$) representing quantum data defined in the Hilbert space; in other words, $\mathbb{W} \subset \mathbb{C}^n$. In this study, MRI data of AD patients, after being converted into multidimensional NumPy arrays, were represented in the Hilbert space using the amplitude encoding technique, one of the most efficient feature mapping approaches in terms of qubit efficiency [92,93]. Moreover, this technique provides high efficiency in terms of memory usage [94,95]. Thus, an $N$-dimensional feature vector is encoded into the superposition amplitudes of quantum states using $log_2 N$ qubits, where $log_2 N$ is equal to the previously defined $n$. In other words, the amplitude encoding method expresses the normalized classical feature vectors ($x = (x_0, x_1, x_2, \dots \dots, x_{2^n-1})$, as the amplitudes of a quantum state:

$$|\psi\rangle = x_0|000 \dots 0\rangle + x_1|00 \dots 1\rangle + \cdots + x_{2^n-1}|11 \dots 1\rangle \quad (1)$$

thereby representing each classical feature component as a corresponding quantum state amplitude.

In this study, the preprocessed AD MR images are expressed in dimensions of 100 x 100 x 3, resulting in a flattened input feature vector of size $N = 30{,}000$. To map these features into the quantum Hilbert space via amplitude encoding using $n = 15$ qubits ($2^{15} = 32{,}768$ state vector amplitudes) the original feature vector $x = (x_0, x_1, \dots, x_{29999})$ was padded with 2,768 zero elements to match the required power-of-two dimension ($2^n$). Consequently, the expanded normalized feature vector $x = (x_0, x_1, \dots, x_{29999}, 0_1, 0_2 \dots, 0_{2768})$ was mapped to the superposition amplitudes of the 15-qubit quantum state.

Here, the class labels corresponding to the ADs are denoted by $y^d$, which are encoded into a one-hot vector representation for three classes using the *to_categorical* function from the Keras library. For instance, in the case of three classes, these labels are represented as [1,0,0], [0,1,0] and [0,0,1], respectively. In this study, only the input data denoted by $x^d$ were mapped from the classical vector space into the quantum Hilbert space, whereas the class labels ($y^d$) remained in the classical space.

### *2.3.2. Parallel Parameterized Quantum Circuits (Parallel PQCs)*

In this part of the study, in order to learn important and distinctive features such as patterns and correlations from data expressed in the form of quantum states, the Parallel PQCs approach is proposed, which operates analogously to the principle of model parallelism in classical computing. Fundamentally, a PQC (also referred to as a trial wave function or an ansatz) is a quantum circuit model that transforms one or more input quantum states into another quantum state by applying one or more tunable single-qubit parameterized gates and non-parameterized multi-qubit quantum gates to the input state [56,81,96–99]. In other words, the PQC model is a structure designed with an architecture similar to classical neural networks, consisting of unitary, adjustable parametric single-qubit gates and/or multi-qubit quantum gates with different entanglement topologies (such as star, ring, or all-to-all), and composed of one or more layers, but defined within the scope of quantum algorithms and equipped with the principles of quantum information processing.

In general, a PQC model can be expressed as

$$W(\theta) = U_{ent}\, U_{rot}^{L}(\theta_L) \dots\ U_{ent}\, U_{rot}^{2}(\theta_2)\, U_{ent}\, U_{rot}^{1}(\theta_1), \tag{2}$$

where $U_{rot}^{l}(\theta_l)$ represents the parameterized rotation block in the $l\text{-}th$ layer, $U_{ent}$ denotes the entanglement block applied after each layer, and the total number of layers is $L$. This expression can equivalently be written as

$$W(\theta) = \prod_{l=1}^{L} \left( U_{ent} U_{rot}^{l}(\theta_l) \right). \tag{3}$$

Currently, due to their powerful expressive capabilities, PQCs demonstrate significant potential in supervised and unsupervised quantum ML, quantum chemistry, and in addressing problems that are challenging for classical algorithms, and they are frequently utilized in the near-term quantum computing domain [100–102]. In this study, two different circuit architectures are defined in the proposed parallel PQCs for learning important patterns and features such as brain atrophy, gray matter loss, and ventricle enlargement in the MR images of AD, and each circuit consists of 20 layers. Moreover, each layer includes parametric single-qubit rotations and multi-qubit entanglement operations. Both PQC architectures initialize the superposition state at the beginning of each layer by applying $Hadamard$ gates to all qubit states. In the subsequent process, in the first PQC circuit, generalized Euler rotation gates ($U3$), $R_x$, $R_y$ parametric quantum gates are applied, followed sequentially by unparametrized fixed Controlled-NOT ($CNOT\ or\ CX$), Toffoli ($CCNOT$), and $CNOT$ gates.

The first $CNOT$ two-qubit quantum gate establishes a "dense all-to-all entanglement" by creating entanglement among all qubits, whereas the second $CNOT$ generates a circular entanglement pattern, referred to as "cyclically repeated entanglement" or "ring entanglement." Thus, while the CNOT gate enables both global and local entanglement, the $CCNOT$ gate couples neighboring triplets, thereby facilitating the formation of more complex correlations within the circuit.

In the first PQC architecture, the parametrized $l\text{-}th$ layer rotational sub-block is defined as

$$U_{rot_1}^{l}(\theta_l) = \prod_{i=0}^{n-1} \left[ R_{y_i}(\theta_{l,i,0})\, R_{x_i}(\theta_{l,i,0})\, U3_i(\theta_{l,i,0}, \theta_{l,i,1}, \theta_{l,i,2})\, H_i \right] \tag{4}$$

where

$$U3(\theta_0, \theta_1, \theta_2) = \begin{bmatrix} \cos\left(\frac{\theta_0}{2}\right) & -e^{i\theta_2}\sin\left(\frac{\theta_0}{2}\right) \\ e^{i\theta_1}\sin\left(\frac{\theta_0}{2}\right) & e^{i(\theta_1+\theta_2)}\cos\left(\frac{\theta_0}{2}\right) \end{bmatrix} \tag{5}$$

is defined as above. Here, $H_i$ denotes the $Hadamard$ gate, $U3_i$ is the Euler rotation gate, $R_{x_i}$ and $R_{y_i}$, represent rotations around the $X$ and $Y$ axes of the Bloch sphere, and $\theta_l = \{\theta_{l,i,k}\}$, represents the learnable parameters of the $l$-th layer. In this architecture, for both learning and optimization processes, the learnable parameters in the $R_x$ and $R_y$ rotational gates are shared by employing a "parameter-sharing approach". Thus, while reducing the number of learnable parameters, the risk of overfitting was mitigated, and the optimization process was accelerated.

The entanglement block of the first PQC architecture is defined as

$$U_{ent_1} = \prod_{i=0}^{n-1} CNOT_{i,\ (i+1)\ mod\ n} \prod_{i=0}^{n-1} CCNOT_{i,\ (i+1)\ mod\ n,\ (i+2)\ mod\ n} \prod_{i=0}^{n-2} \prod_{j=i+1}^{n-1} CNOT_{i \to j} \quad (6)$$

The first PQC architecture in its final form is expressed as follows:

$$W_1(\theta) = \prod_{l=1}^{L} \left[ U_{ent_1} U_{rot_1}^{l}(\theta_l) \right] \quad (7)$$

In this study, the number of layers is set to L=20.

The second PQC model, similar in architecture to the first PQC model, consists of gates that combine parametric single-qubit transformations with multi-qubit entangling operations. In the second PQC architecture, after applying a $Hadamard$ gate to all qubits at the beginning of each layer, each qubit is subsequently acted upon by the Generalized Euler Rotation Gate ($U3$), as well as the parameterized quantum gates $R_x$ and $R_y$. Following these, the Controlled-Y gate ($CY$) and the Toffoli gate ($CCNOT$) are applied as fixed, non-parameterized gates. In this entanglement structure, the two-qubit $CY$ gate enables global entanglement through "dense all-to-all entanglement," while the $CCNOT$ gate, as in the first PQC architecture, couples neighboring triplets to establish strong correlations.

In the second PQC architecture,

the parametrized $l$-th layer rotational sub-block is defined as

$$U_{rot_2}^{l}(\theta_l) = \prod_{i=0}^{n-1} \left[ R_{y_i}(\theta_{l,i,0})\, R_{x_i}(\theta_{l,i,0})\, U3_i(\theta_{l,i,0}, \theta_{l,i,1}, \theta_{l,i,2})\, H_i \right] \quad (8)$$

The entanglement block of this circuit, $U_{ent_2}$, is defined as

$$U_{ent_2} = \prod_{i=0}^{n-1} CCNOT_{i,\ (i+1)\ mod\ n,\ (i+2)\ mod\ n} \prod_{i=0}^{n-2} \prod_{j=i+1}^{n-1} CY_{i \to j} \quad (9)$$

The second PQC architecture in its final form is expressed as follows:

$$W_2(\theta) = \prod_{l=1}^{L} \left[ U_{ent_2} U_{rot_2}^{l}(\theta_l) \right] \quad (10)$$

As in the first PQC architecture, the second architecture employs a total of $L = 20$ layers. The features and matrix representations of the parameterized single-qubit gates and non-parameterized fixed multi-qubit gates employed in this study are presented in **Table 1**. The circuit structures corresponding to the first and second PQC architectures are illustrated in **Figure 4**.

**Table 1.** Circuit representations, qubit counts, and matrix definitions of the parameterized single-qubit quantum gates ($R_x$, $R_y$, $U3$) and non-parameterized multi-qubit gates (CNOT, CY, CCNOT) implemented on the PennyLane platform.

| **Quantum Gate** | **Number of Qubits** | **Quantum Circuit Structure in PennyLane** | **Matrix Representation** |
|---|---|---|---|
| $R_x$ | Single-Qubit | 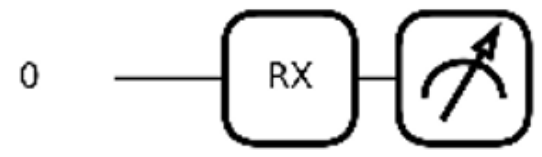 | $R_x(\theta) = e^{-iX\frac{\theta}{2}} = \begin{bmatrix} \cos\left(\frac{\theta}{2}\right) & -\sin\left(\frac{\theta}{2}\right) i \\ -\sin\left(\frac{\theta}{2}\right) i & \cos\left(\frac{\theta}{2}\right) \end{bmatrix}$ |
| $R_y$ | Single-Qubit | 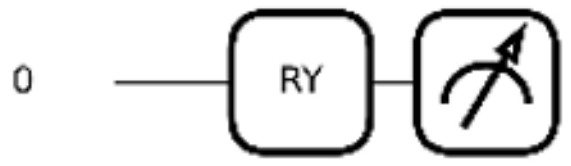 | $R_y(\theta) = e^{-iY\frac{\theta}{2}} = \begin{bmatrix} \cos\left(\frac{\theta}{2}\right) & -\sin\left(\frac{\theta}{2}\right) \\ \sin\left(\frac{\theta}{2}\right) & \cos\left(\frac{\theta}{2}\right) \end{bmatrix}$ |
| *Generalized Euler Rotation Gates* $(U3)$ | Single-Qubit | 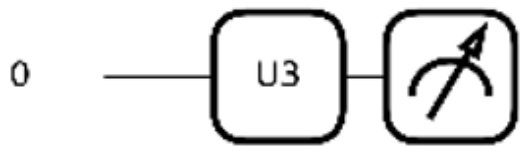 | $U3(\theta_0, \theta_1, \theta_2) = \begin{bmatrix} \cos\left(\frac{\theta_0}{2}\right) & -e^{i\theta_2} \sin\left(\frac{\theta_0}{2}\right) \\ e^{i\theta_1} \sin\left(\frac{\theta_0}{2}\right) & e^{i(\theta_1+\theta_2)} \cos\left(\frac{\theta_0}{2}\right) \end{bmatrix}$ |
| *Controlled-NOT (CNOT or CX)* | Two-Qubit | 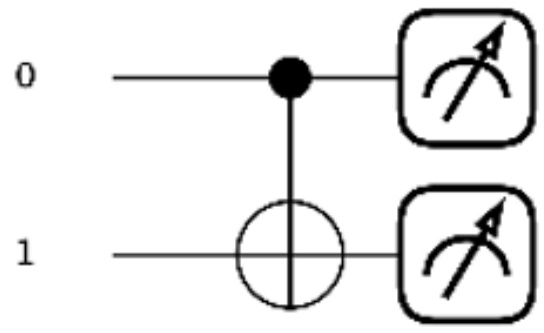 | $\begin{bmatrix} 1 & 0 & 0 & 0 \\ 0 & 1 & 0 & 0 \\ 0 & 0 & 0 & 1 \\ 0 & 0 & 1 & 0 \end{bmatrix}$ |
| *Controlled-Y (CY)* | Two-Qubit | 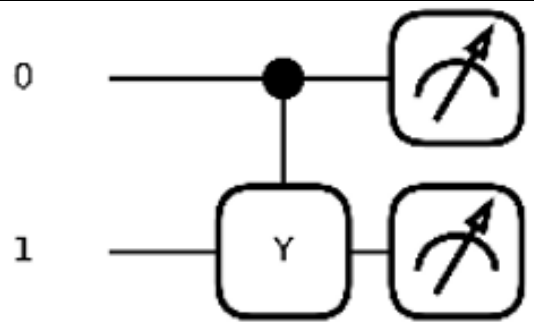 | $\begin{bmatrix} 1 & 0 & 0 & 0 \\ 0 & 1 & 0 & 0 \\ 0 & 0 & 0 & -i \\ 0 & 0 & i & 0 \end{bmatrix}$ |
| *Toffoli (CCNOT)* | Three-Qubit | 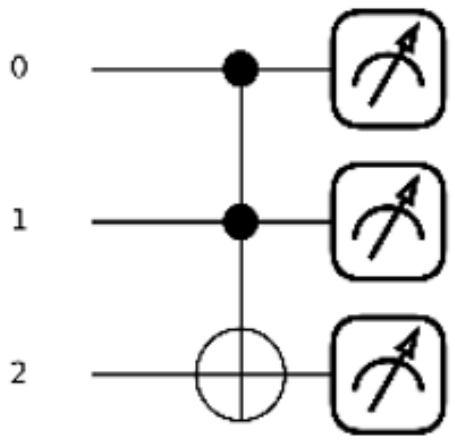 | $\begin{bmatrix} 1 & 0 & 0 & 0 & 0 & 0 & 0 & 0 \\ 0 & 1 & 0 & 0 & 0 & 0 & 0 & 0 \\ 0 & 0 & 1 & 0 & 0 & 0 & 0 & 0 \\ 0 & 0 & 0 & 1 & 0 & 0 & 0 & 0 \\ 0 & 0 & 0 & 0 & 1 & 0 & 0 & 0 \\ 0 & 0 & 0 & 0 & 0 & 1 & 0 & 0 \\ 0 & 0 & 0 & 0 & 0 & 0 & 0 & 1 \\ 0 & 0 & 0 & 0 & 0 & 0 & 1 & 0 \end{bmatrix}$ |

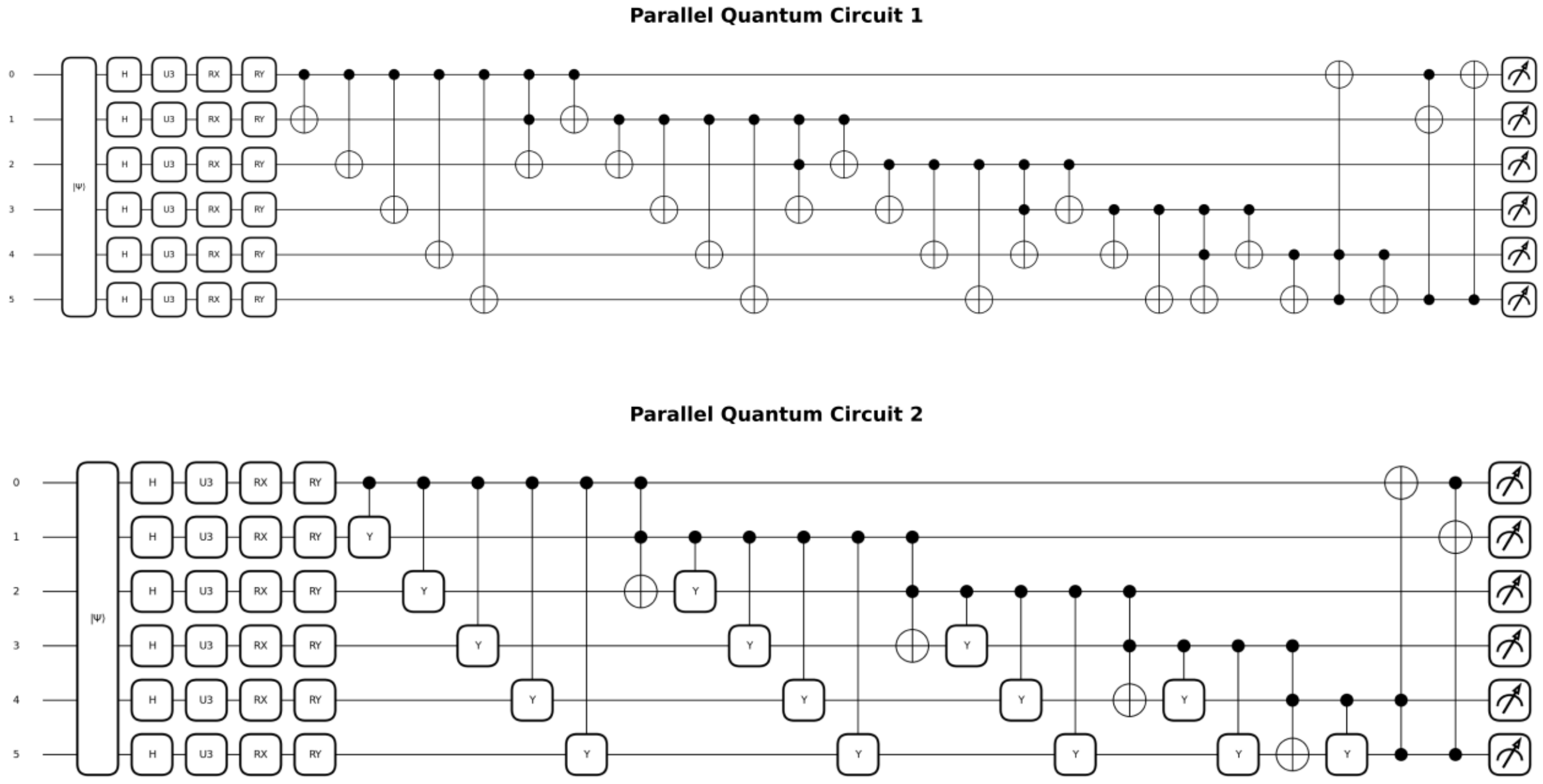

**Figure 4.** Quantum circuit diagrams of the first and second parallel PQC structures defined within the proposed QBPM architecture. Both circuits are executed in parallel on the same quantum simulator. The first circuit comprises $U3$, $R_x$, $R_y$, and $CNOT$, $CCNOT$ quantum gates, whereas the second circuit, in addition to the same parametric structure, includes $CY$ and $CCNOT$ entangling gates. Because of space restrictions, only the initial layer is presented, though the full architecture includes 20 layers.

### 2.3.3. *Measurement*

Both PQC models operate in parallel on the quantum states obtained from data of Alzheimer's patients, learning meaningful features and patterns. Subsequently, measurement operations are performed on all qubits in the Pauli-Z basis for both circuit structures, allowing class prediction values to be obtained separately for each of the two circuit models.

The final quantum transformation resulting from combining the first PQC model ($W_1(\theta)$) with amplitude encoding, denoted as $U_1(\theta, x)$, and the final quantum transformation resulting from combining the second PQC model ($W_2(\theta)$) with amplitude encoding, denoted as $U_2(\theta, x)$, yield classical prediction values when measured individually in the Pauli-Z basis:

$$
\begin{aligned}
logit_k &= \left\langle 0^{\otimes n} \middle| U_1(\theta, x)^{\dagger} Z_k U_1(\theta, x) \middle| 0^{\otimes n} \right\rangle \\
logit_{k'} &= \left\langle 0^{\otimes n} \middle| U_2(\theta, x)^{\dagger} Z_{k'} U_2(\theta, x) \middle| 0^{\otimes n} \right\rangle
\end{aligned} \tag{11}
$$

where $Z_k$ and $Z_{k'}$ represent the Pauli-Z operators acting on qubits $k$ and $k'$, respectively. The measurement output components obtained from these operations, $logit_k$ and $logit_{k'}$, take values within the interval [-1, 1] and constitute classical output vectors composed of the measurement expectations of all qubits, with $logit_k,\ logit_{k'} \in \mathbb{R}^n$. In the next step, the obtained prediction output vectors are combined using the $concat$ function to yield the final output. This operation can be expressed as

$$logit_{concat} = concat\ (logit_k, logit_{k'}), \tag{12}$$

where $logit_{concat} \in \mathbb{R}^{2n}$.

This procedure transforms the measurement outputs obtained from the two separate parallel PQC architectures into a joint representation vector, thereby enabling a "unified decision-making mechanism".

Classical Computing

### *2.3.4. Softmax, Loss Function, and Optimization*

In the quantum computing section of the proposed QBPM architecture, after feature mapping and parallel PQC operations, a measurement in the Pauli-Z basis was performed, and the average of the obtained results was taken to calculate the expected value. In a quantum computing model, the expectation value of an observable is the statistical mean of results obtained by preparing and measuring the system in the same initial state multiple times. In this way, a result close to the theoretical expectation value is obtained. The calculated expected value represents the prediction score of the model and lies within the interval:

$$\langle Z \rangle \in [-1, 1] \tag{13}$$

The mathematical correspondences of the operations performed up to this point are expressed in Equations 4–12. In the next stage, in order to interpret the outputs obtained from the quantum model and to derive probability distributions from the prediction scores, the softmax function was employed in the classical part of the model. Although the measurements were performed over all qubits, the final obtained logit vector (from Equation 12) was normalized and processed, and three components of this vector were selected to represent three different Alzheimer's classes. These three logit values were then combined with a learnable three-dimensional bias vector ($bias \in \mathbb{R}^3$) to provide additional flexibility in class predictions. In other words, this step is defined as

$$\begin{gathered} |\psi_1\rangle = U_1(\theta, x)|0^{\otimes n}\rangle,\ logit_k = \langle\psi_1|Z_k|\psi_1\rangle, \\ |\psi_2\rangle = U_2(\theta, x)|0^{\otimes n}\rangle,\ logit_{k'} = \langle\psi_2|Z_{k'}|\psi_2\rangle. \end{gathered} \tag{14}$$

In this case,

$$\begin{gathered} logit_{concat} = concat(logit_k, logit_{k'}) = \\ concat(\langle\psi_1|Z_k|\psi_1\rangle, \langle\psi_2|Z_{k'}|\psi_2\rangle) \in \mathbb{R}^{2n} \end{gathered} \tag{15}$$

Subsequently, these values were input into the softmax function, thereby obtaining probability distributions for each class.

Here, the logit vector,

$$\begin{gathered} logit = \left[logit_{concat_1}, logit_{concat_2}, logit_{concat_3}\right]^T + bias \in \mathbb{R}^3 \\ bias = [bias_1, bias_2, bias_3]^T \in \mathbb{R}^3 \end{gathered} \tag{16}$$

is provided as input to the softmax function to obtain class probabilities:

$$p_i = \frac{\exp(logit_i)}{\sum_{j=1}^{3} \exp(logit_j)} \tag{17}$$

$$i = 1,2,3$$

At this stage, to quantify the difference between the class labels predicted by the model ($\widehat{y_i}$) and the true Alzheimer's class labels ($y_i$), the probability values obtained from the softmax function ($p_i$) are employed within the cross-entropy loss function ($C(\theta)$):

$$C(\theta) = -\sum_{i=1}^{3} y_i ln(p_i) \tag{18}$$

In order for the proposed QBPM architecture to correctly predict the labels of Alzheimer's classes, it is necessary to determine the optimal $\theta$ parameters that minimize the loss function. This process is carried out using a hybrid training approach that combines classical and quantum devices; the classical device is responsible for the computation of the loss function and the optimization steps, while the quantum device is responsible for calculating the gradients at this stage. In optimization procedures, gradient-based strategies are commonly employed, and in this study, the classical Adam optimizer was used. Adaptive Moment Estimation (or Adam) computes a step-dependent learning rate for each parameter by utilizing estimates of the first and second moments of the gradients [103]. This can be expressed as follows:

$$\theta^{(t+1)} = \theta^{(t)} - \frac{\eta^{(t+1)}}{\sqrt{v^{(t+1)}} + \varepsilon} m^{(t+1)}$$

$$m^{(t+1)} = \beta_1 m^{(t)} + (1-\beta_1)\nabla C(\theta^{(t)}) \tag{19}$$

$$v^{(t+1)} = \beta_2 v^{(t)} + (1-\beta_2)\left(\nabla C(\theta^{(t)})\right)^{\odot 2}$$

$$\eta^{(t+1)} = \eta . \frac{\sqrt{1-{\beta_2}^{t+1}}}{1-{\beta_1}^{t+1}}$$

Here, $\left(\nabla C(\theta^{(t)})\right)^{\odot 2}$ denotes the element-wise squaring operation. In this study, since the derivative of the loss function is reduced to the gradient of the expectation value over the predictions, the parameter shift rule method was employed at this stage. Thus, the gradient of the quantum computation can be calculated via the derivative of the expectation value.

Quantum Gradient Computation Using the Parameter-Shift Rule

The parameter-shift rule is a method employed on quantum hardware for the efficient computation of partial derivatives of PQCs. With this method, the gradient of the expectation value with respect to gate parameters is calculated, and no additional ancilla qubit is required at this stage. Moreover, a fundamental requirement for the applicability of the parameter-shift rule is that the gate parameters possess two distinct eigenvalues [104,105].

The expectation value with respect to the observable $B$, for a quantum state generated from the initial state $|\psi\rangle$ by a parameterized quantum circuit $U_G(\theta)$ characterized by the tunable parameter vector $\theta$, is calculated as:

$$f(\theta) = \langle\psi|U_G(\theta)^\dagger B\, U_G(\theta)|\psi\rangle \tag{20}$$

where $f(\theta)$ represents the objective function of the quantum circuit. The parameterized circuit $U_G(\theta)$ can be expressed as

$$U_G(\theta) = e^{-ia\theta G}, \tag{21}$$

where $G$ is a Hermitian operator (or Hermitian generator) and $a$ is a real constant. One of the most important conditions is that the Hermitian operator $G$ possesses two distinct eigenvalues. When $G$ has two distinct eigenvalues (e.g., $e_{0,}$ $e_1$), shifting these eigenvalues by $\pm r$ affects the global phase; however, this phase is not observable in quantum systems and does not alter the physical outcomes (e.g., measurement results). In this context, the parameter-shift rule states that the derivative of the expectation value with respect to $\theta$ for a Hermitian operator $G$ with two distinct eigenvalues is proportional to the difference between the expectation values of two circuits with the parameter shifted by a specific amount. This can be expressed as

$$\frac{d}{d\theta} f(\theta) = r\left[f\left(\theta + \frac{\pi}{4r}\right) - f\left(\theta - \frac{\pi}{4r}\right)\right] \tag{22}$$

where $r = \frac{a}{2}(e_1 - e_0)$ is the phase constant. More detailed mathematical expressions of the parameter-shift rule can be found in [104]. For Pauli rotation gates, the phase constant can be taken as $r = \frac{1}{2}$.

In our study, this procedure is applied separately to two distinct parallel circuits, and the parameters are updated simultaneously. This is expressed as

$$\begin{aligned} \frac{\partial}{\partial {\theta_i}^{(1)}} \langle\psi_1|Z_k|\psi_1\rangle &= \frac{1}{2}\left[\langle\psi_1|Z_k|\psi_1\rangle_{{\theta_i}^{(1)}+\frac{\pi}{2}} - \langle\psi_1|Z_k|\psi_1\rangle_{{\theta_i}^{(1)}-\frac{\pi}{2}}\right] \\ \frac{\partial}{\partial {\theta_j}^{(2)}} \langle\psi_2|Z_{k'}|\psi_2\rangle &= \frac{1}{2}\left[\langle\psi_2|Z_{k'}|\psi_2\rangle_{{\theta_j}^{(2)}+\frac{\pi}{2}} - \langle\psi_2|Z_{k'}|\psi_2\rangle_{{\theta_j}^{(2)}-\frac{\pi}{2}}\right] \end{aligned} \tag{23}$$

where ${\theta_i}^{(1)}$ and ${\theta_j}^{(2)}$ are the tunable parameters corresponding to the first and second quantum circuits, respectively.

### *2.4. Barren Plateau Analysis*

To evaluate the trainability and theoretical scalability of the proposed QBPM architecture as the number of qubits increases, the variance of the cost function gradient was analyzed as a function of qubit count. Representing this variance over the parameter distribution $\theta$ as $\mathrm{Var}_\theta[\partial_i C]$, the mathematical condition for a barren plateau is defined as:

$$Var_\theta[\partial_i C] \leq F(n), \qquad F(n) \in \mathcal{O}\left(\frac{1}{b^n}\right) \tag{24}$$

Here, $(b > 1)$. Combining this formulation with Chebyshev's inequality bounds the probability of the cost function partial derivative deviating from zero:

$$Pr(|\partial_i C| \geq c) \leq \frac{Var_\theta[\partial_i C]}{c^2} \leq \frac{F(n)}{c^2} \tag{25}$$

Under this theoretical framework, the existence of a barren plateau within the optimization landscape was evaluated by tracking the gradient variance scaling. Specifically, the variation in cost function gradient variance with respect to increasing qubit capacity was conducted both prior to (random initialization) and post-training.

### *2.5. Evaluation Metrics*

In this study, the classification performance of the proposed QBPM architecture in distinguishing between the stages of AD was evaluated using metrics such as average training loss, validation loss, training accuracy, and validation accuracy, with the corresponding variation graphs presented for each iteration. Additionally, to evaluate the overall classification performance, statistical metrics including precision, recall, and $F_1$-score were computed for each class individually.

## 3. Results

In this section, the performance of the proposed QBPM architecture in distinguishing AD stages was evaluated both on the original dataset and on an external dataset used for model validation, and the obtained results are presented in detail. This section is organized into seven subsections:

- Results on the original dataset,
- Results obtained when Gaussian noise is added to the images in the original dataset,
- Results obtained when artificial gate-level noise is introduced to the quantum circuit on the original dataset,
- Results on the external validation dataset,
- Results obtained when Gaussian noise is added to the images in the external validation dataset,
- Results obtained when artificial gate-level noise is introduced to the quantum circuit on the external validation dataset,
- Results of classical baseline models for comparison.

In the study, the learning rate for the Adam optimizer was set to 0.1, while the regularization parameter, lambda was set to 0.01. To prevent overfitting in each computational process, the number of epochs was set to 15, and the batch size was 8. The corresponding average training loss, validation loss, training accuracy, and validation accuracy values were recorded for each iteration (epoch). Additionally, for each process, the number of qubits was set to 15, and the number of circuit layers was set to 20.

Within the scope of this research, the quantum circuits were executed on the default.qubit state-vector simulator provided by the PennyLane 0.35.1 framework. The results reported are based on computations performed on the Perlmutter supercomputer at NERSC. For training the DL models, the Keras library with TensorFlow as the backend was utilized. The machine used for carrying out these processes was equipped with a 12th Gen Intel(R) Core(TM) i5 processor and 8.0 GB of RAM.

### *3.1. Results on the Original Dataset*

In this section, the performance of the proposed QBPM architecture in distinguishing between the three stages of AD present in the original dataset—very mild dementia, mild dementia, and moderate dementia—was evaluated. The classification performance of the model was analyzed based on the accuracy and loss values obtained throughout the training and validation processes. The average training and validation accuracies were calculated as 0.90 and 0.93, respectively, while the training and validation losses were 1.65 and 1.85. The corresponding accuracy and loss graphs are presented in **Figure 5**. To further examine the classification performance of the model, the resulting confusion matrix is shown in **Figure 6**.

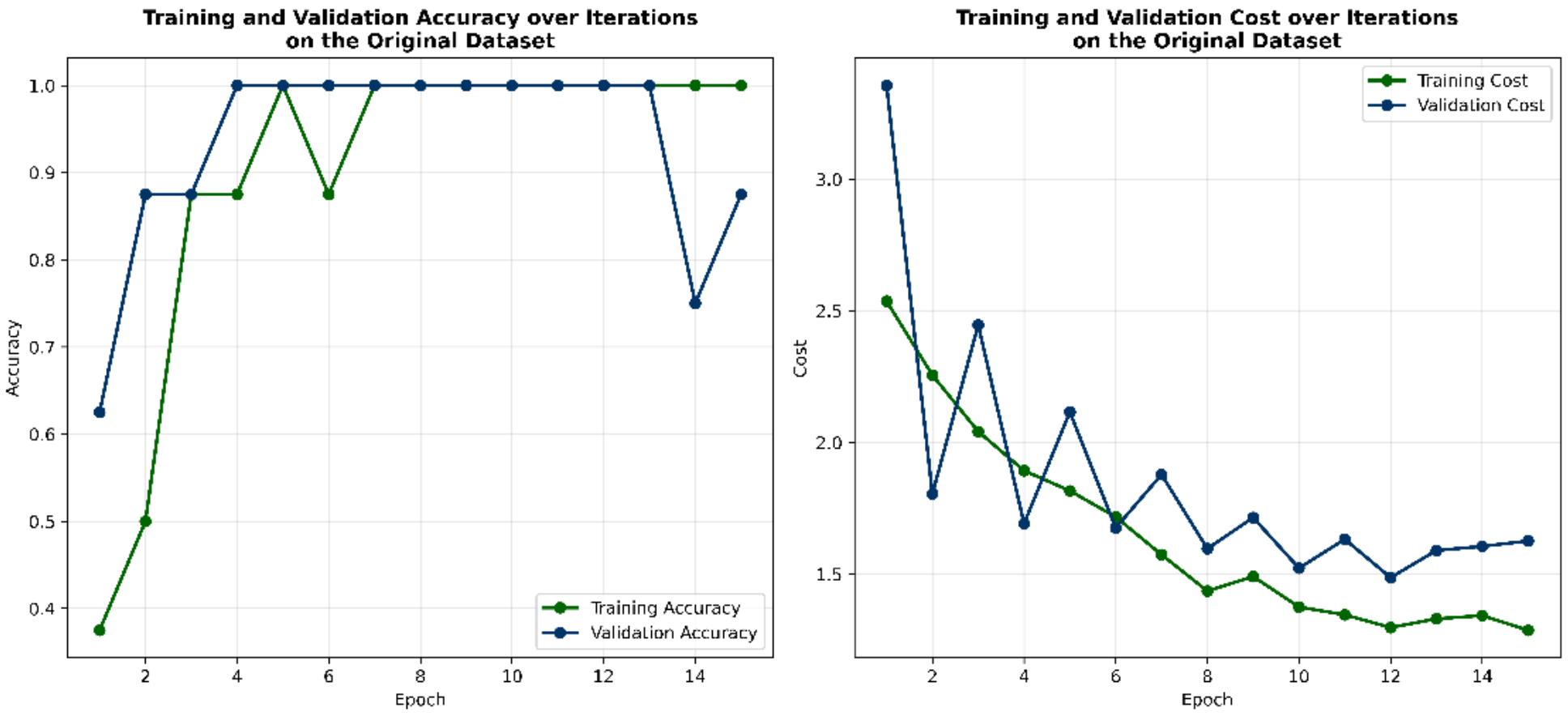

**Figure 5.** Epoch-wise variation of accuracy and loss values during the training and validation phases of the proposed QBPM architecture on the original dataset.

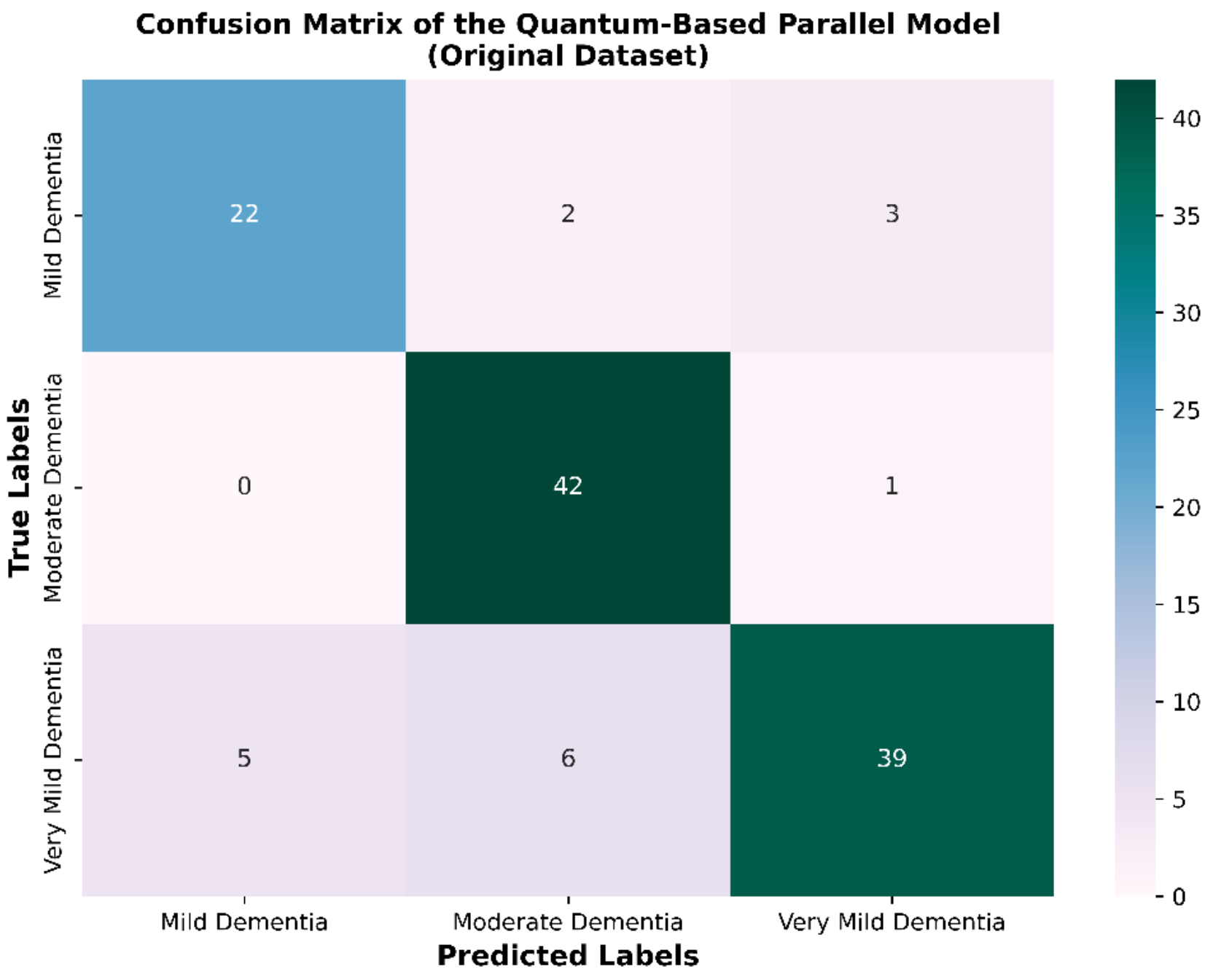

**Figure 6.** Confusion matrix of the classification results obtained using the proposed QBPM architecture on the original dataset.

Furthermore, for the three AD stages, the obtained precision, recall, and $F_1$-score values range from 0.81 to 0.91, 0.78 to 0.98, and 0.81 to 0.90, respectively. Detailed information regarding these values is presented in **Table 2**.

### *3.2. Results obtained when Gaussian noise is added to the images in the original dataset*

In this section, Gaussian (normal) noise was added to the Alzheimer's MR images in order to evaluate whether the proposed QBPM architecture is suitable for real-world conditions, as well as to test its generalization ability and robustness. In this way, the model's resilience against real-world factors—such as variability caused by different scanners, operator errors, and imaging settings—was assessed. To investigate the effect of noisy data on classification performance, random numbers drawn from a Gaussian distribution with a mean of 0 and a standard deviation of 0.01 were added to each pixel. Additionally, a high noise factor of 0.5 was employed. Under these conditions, the obtained average training accuracy, validation accuracy, training loss, and validation loss were 0.87, 0.88, 1.68, and 1.92, respectively. **Figure 7** illustrates the variation of accuracy and loss during the training and validation phases, while **Figure 8** presents the confusion matrix of the model.

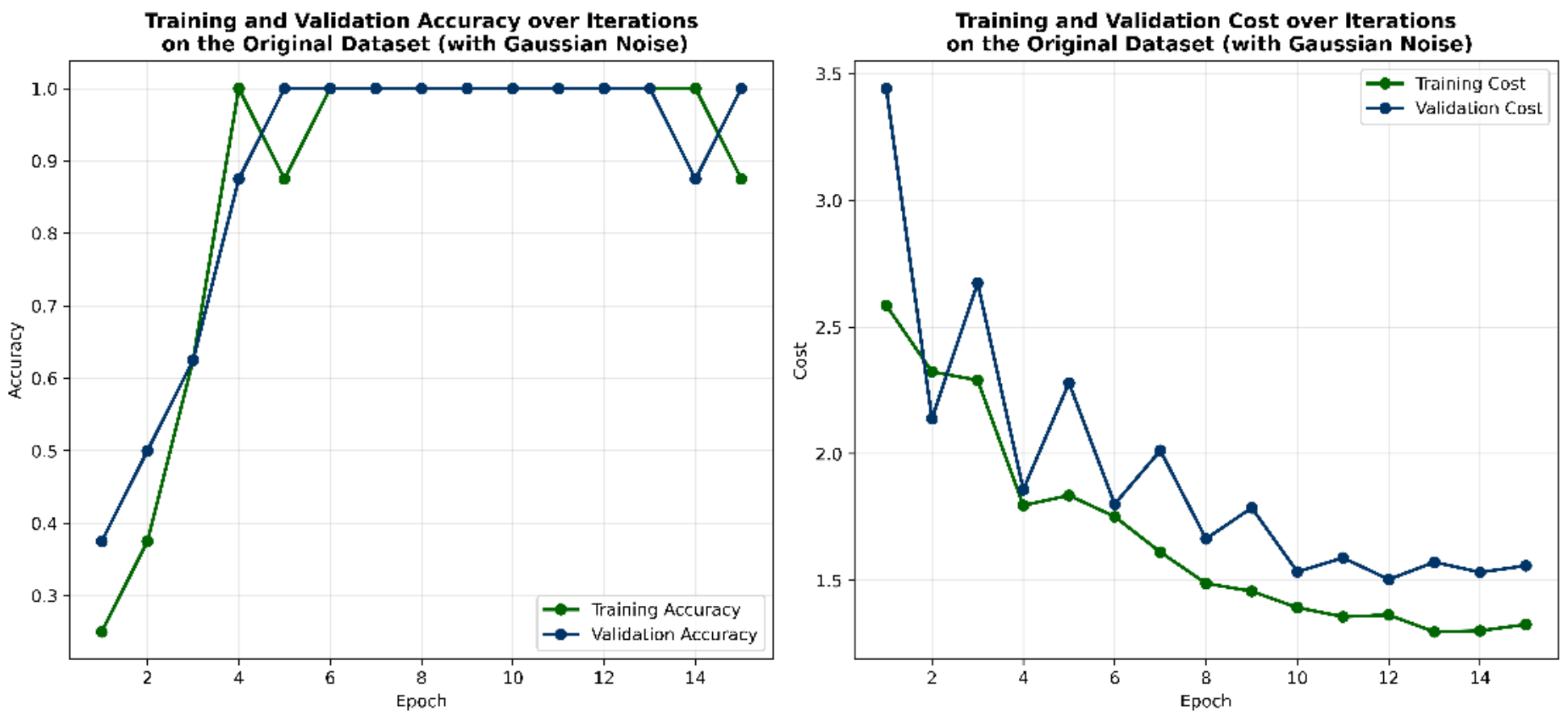

**Figure 7.** Visualization of the training and validation accuracy/loss curves obtained with the proposed QBPM architecture following the addition of Gaussian noise to the MR images in the original dataset.

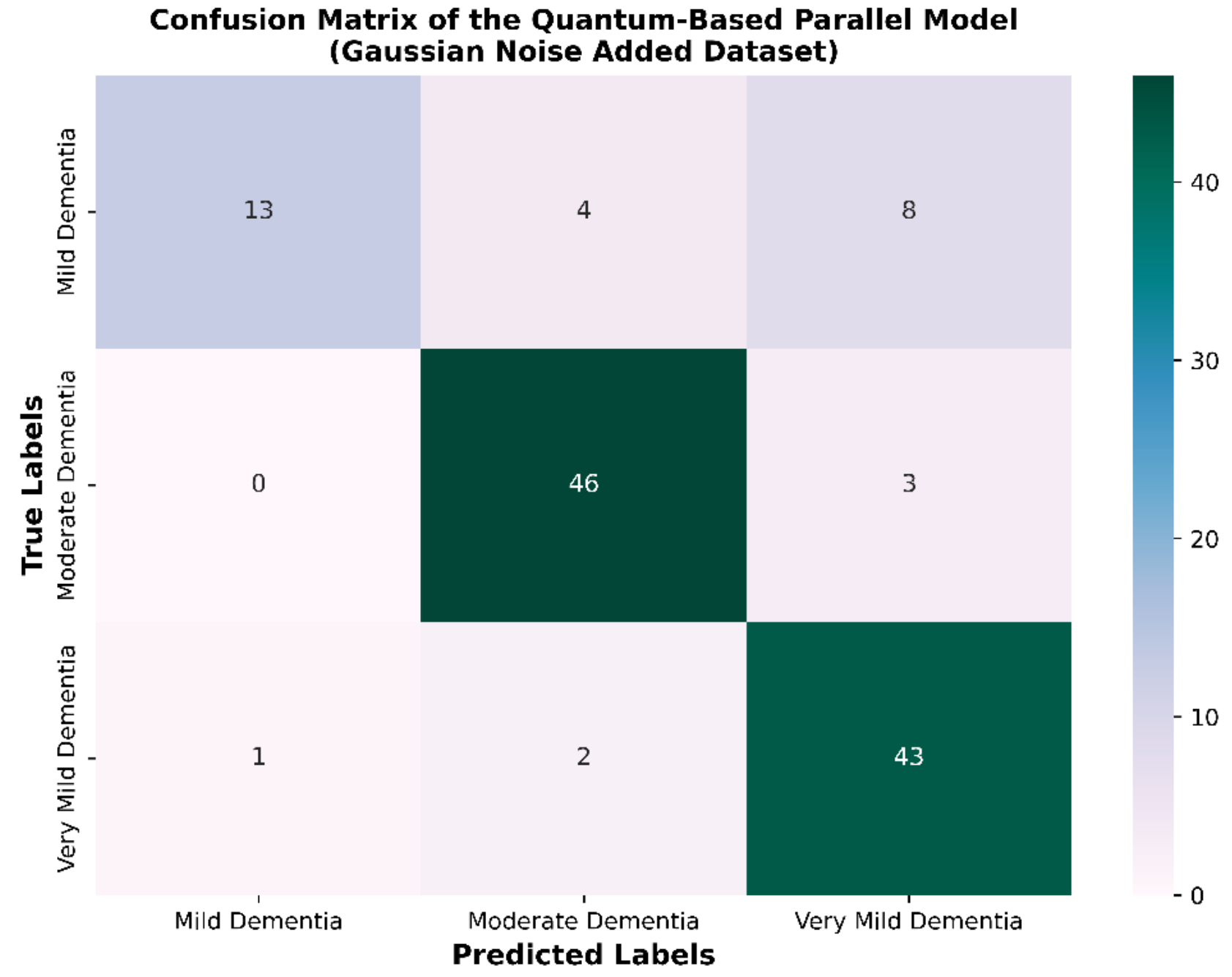

**Figure 8.** Confusion matrix illustrating the classification performance of the QBPM architecture trained on MR images with added Gaussian noise.

In addition, the precision, recall, and $F_1$-score for the three stages vary within the ranges of 0.80–.93, 0.52–0.94, and 0.67–0.91, respectively, with detailed results provided in **Table 2**.

### *3.3. Results obtained when artificial gate-level noise is introduced to the quantum circuit on the original dataset*

In this section of the study, to reflect gate-induced errors occurring in a real quantum computing environment to a certain extent, and to model the impact of gate noise within a quantum simulator, artificial noise was added to the angle parameters of each quantum gate, and the effects of this perturbation on classification performance were examined. For single-qubit quantum gates, namely the $U3$, $R_x$, $R_y$ gates, random numbers drawn from a Gaussian distribution with mean 0 and standard deviation 0.01 were added to their angle values. For multi-qubit quantum gates, specifically $CNOT$, $CY$ and $CCNOT$ instead of Gaussian noise, phase noise was introduced through the $R_z$ gate, thereby attempting to emulate real-world phase shift to a certain degree. As a result, the average accuracy and loss values were obtained as 0.83 and 1.83 for training, and 0.77 and 2.28 for validation. The accuracy and loss curves corresponding to these results are presented in **Figure 9**, while the associated confusion matrix is shown in **Figure 10**.

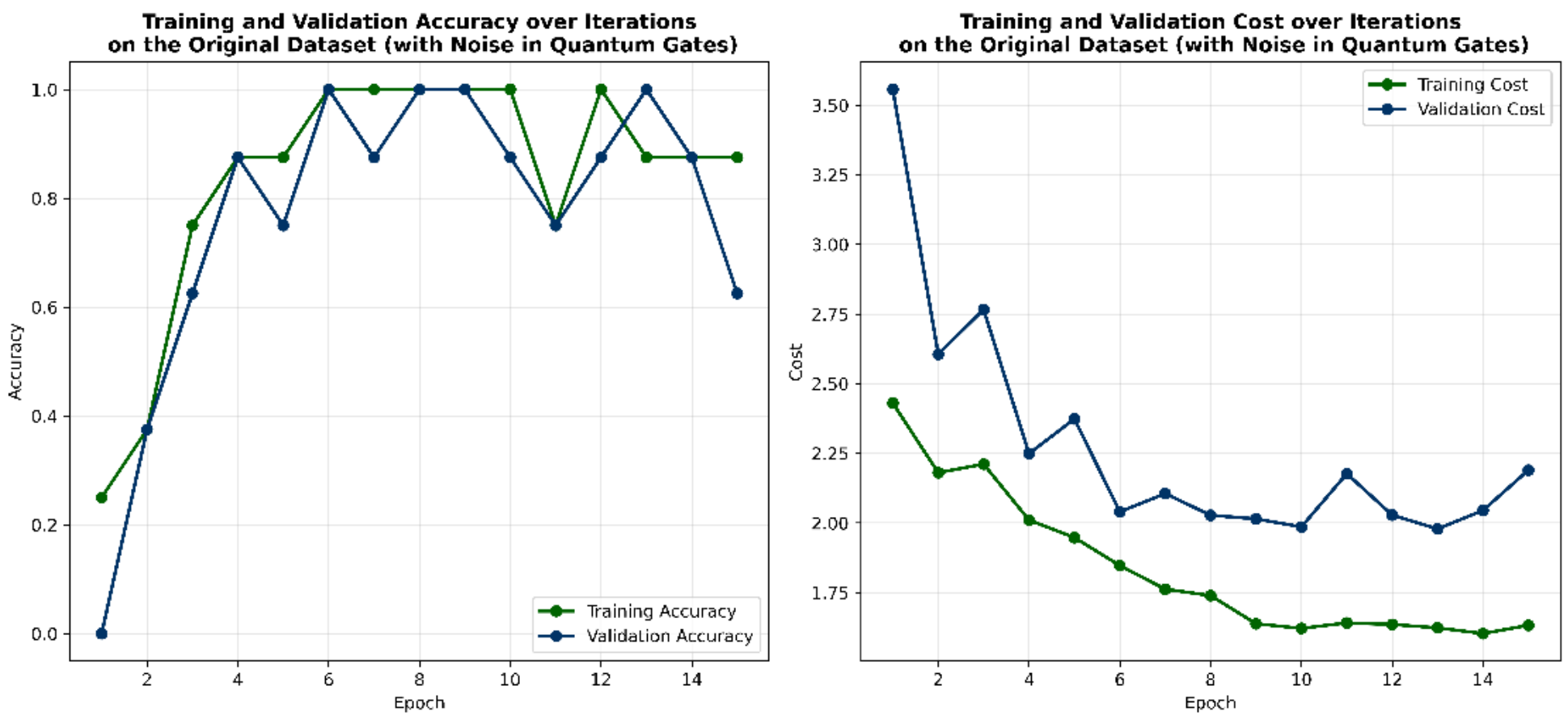

**Figure 9.** Accuracy and loss curves for training and validation on the original dataset after adding artificial noise to the quantum gates within the proposed QBPM architecture.

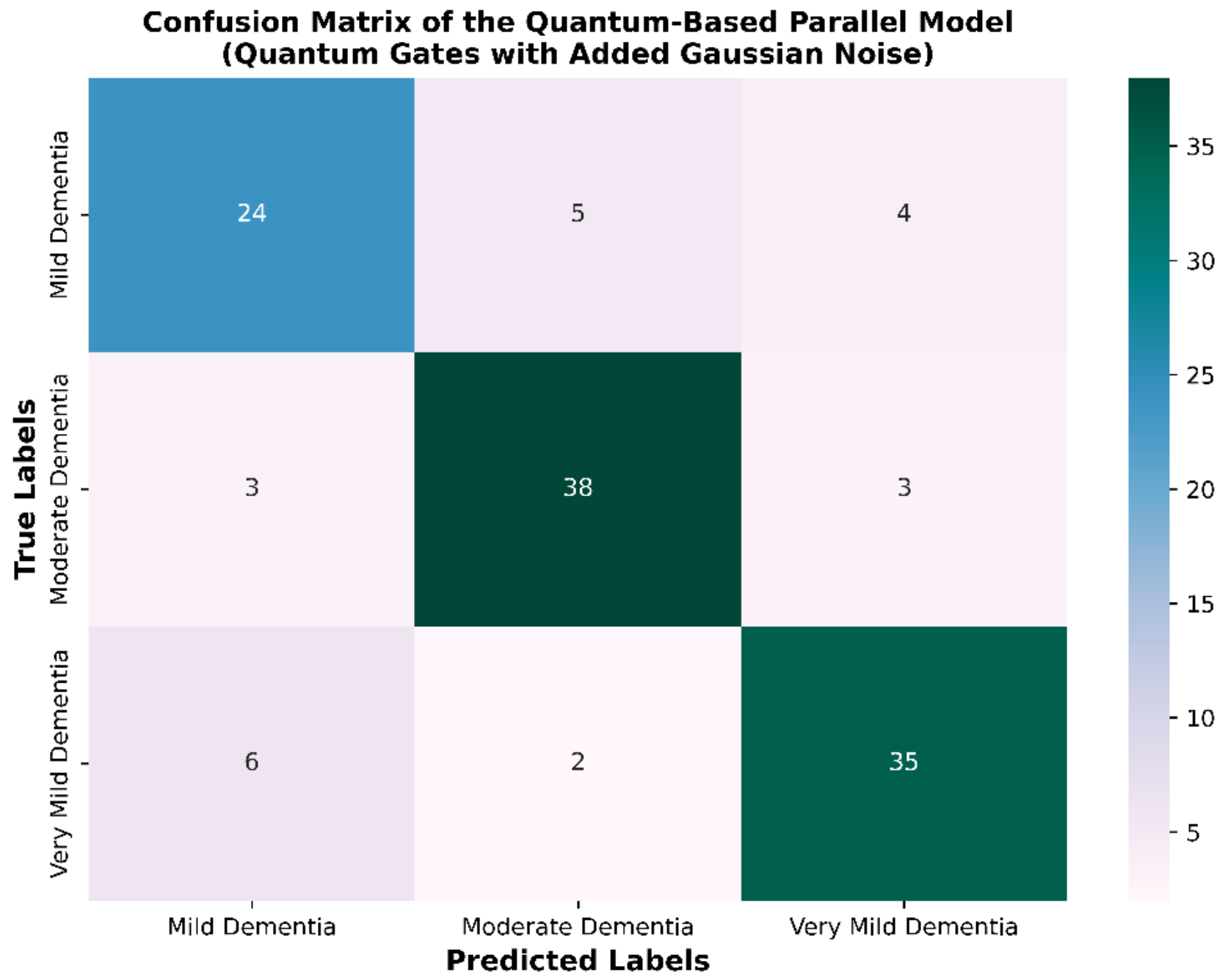

**Figure 10.** Confusion matrix obtained on the original dataset after adding artificial noise to the quantum gates in the proposed QBPM architecture.

Across the three classes of AD, the model achieved precision, recall, and $F_1$-score values ranging from 0.73–0.84, 0.73–0.86, and 0.74–0.85, respectively, with detailed results presented in **Table 2**.

### *3.4. Results on the external validation dataset*

In this part of the study, to assess the robustness and generalization capability of the proposed QBPM architecture, the ADNI: Alzheimer's MRI Classification Dataset, comprising four classes—Cognitively Normal, Early Mild Cognitive Impairment, Late Mild Cognitive Impairment, and Alzheimer's Disease— was employed, and an attempt was made to distinguish between these stages. Upon executing the algorithm, the average training accuracy, validation accuracy, training loss, and validation loss on the external validation dataset were 0.85, 0.81, 1.66, and 1.86, respectively. The corresponding accuracy and loss curves are presented in **Figure 11**, while the associated confusion matrix is shown in **Figure 12**.

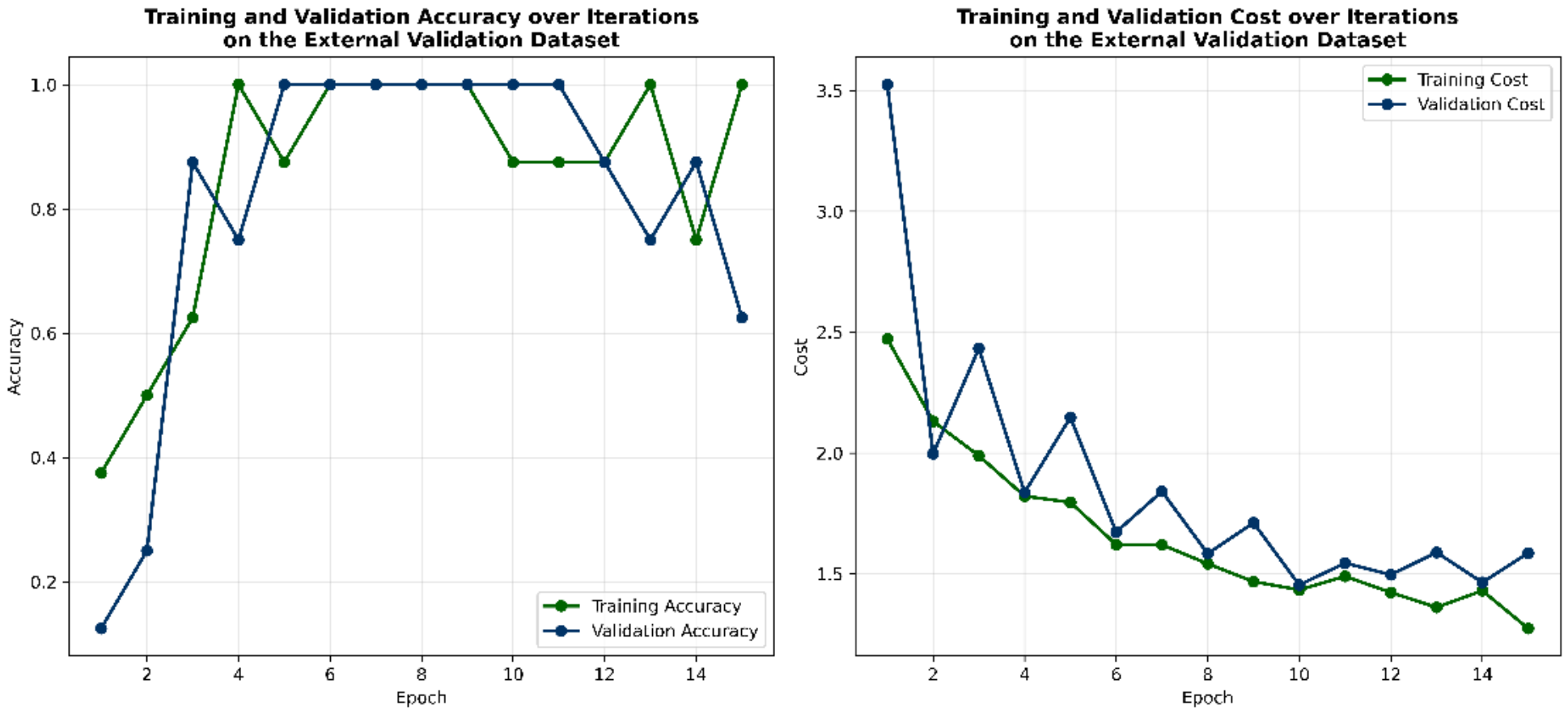

**Figure 11.** Visualization of the training and validation accuracy and loss curves obtained on the external validation dataset using the proposed QBPM architecture.

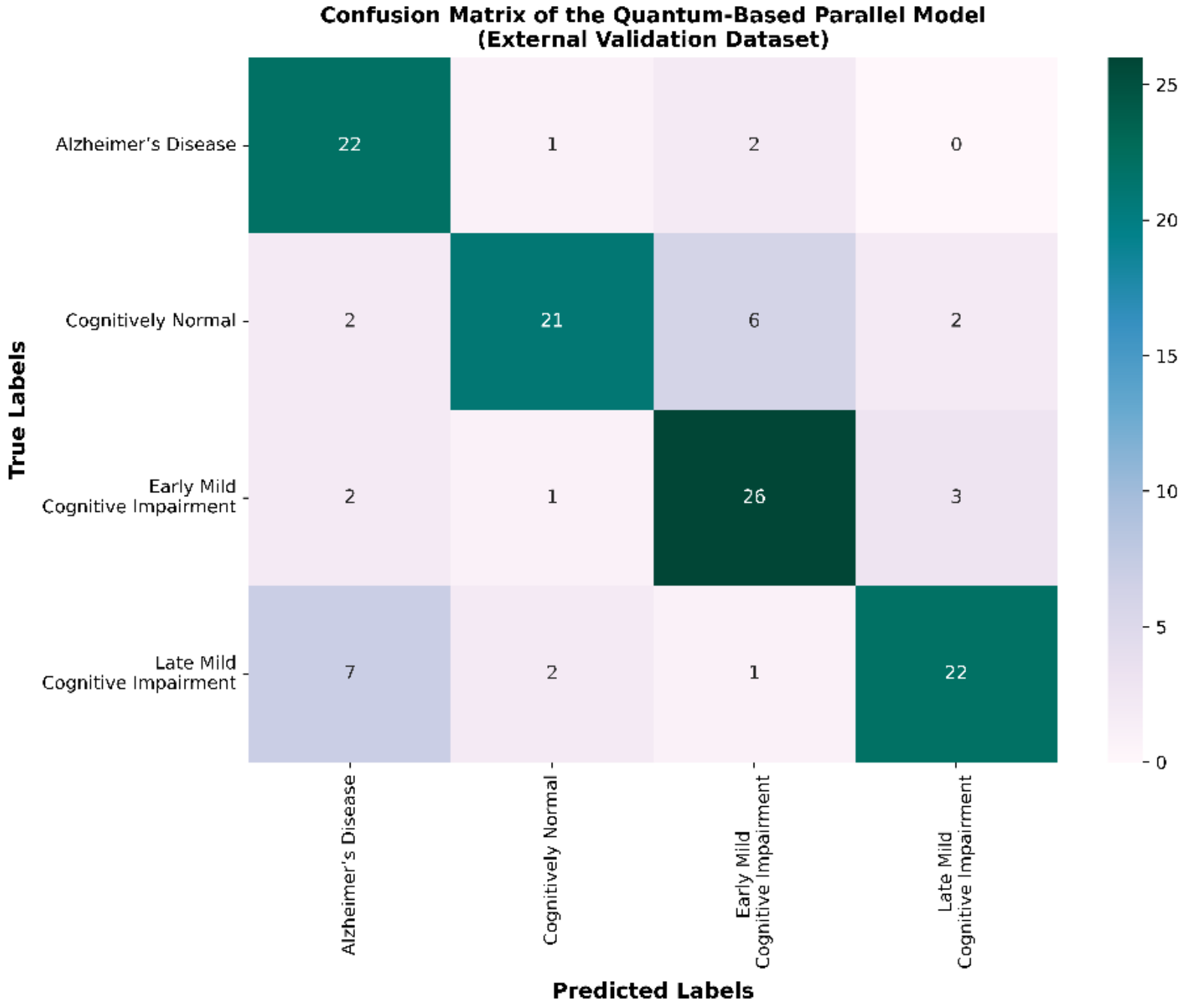

**Figure 12.** Confusion matrix illustrating the classification performance of the QBPM architecture trained on the external validation dataset.

Moreover, for the four classes—Cognitively Normal, Early Mild Cognitive Impairment, Late Mild Cognitive Impairment, and Alzheimer's Disease—the precision, recall, and $F_1$-score range from 0.67 to 0.84, 0.69 to 0.88, and 0.75 to 0.78, respectively, with detailed information provided in **Table 3**.

### *3.5. Results obtained when Gaussian noise is added to the images in the external validation dataset*

In this section, the Gaussian noise added to the Alzheimer's MR images in the original dataset was similarly applied to the MR images in the external validation dataset, and the model's classification performance was evaluated under these conditions. Random numbers drawn from a Gaussian distribution with mean 0 and standard deviation 0.01 were added to each pixel, and the noise factor was set to 0.5. The resulting average training and validation accuracies were 0.68 and 0.60, respectively, while the corresponding training and validation losses were 1.69 and 1.87. The accuracy and loss curves are presented in **Figure 13**, and the associated confusion matrix is shown in **Figure 14**.

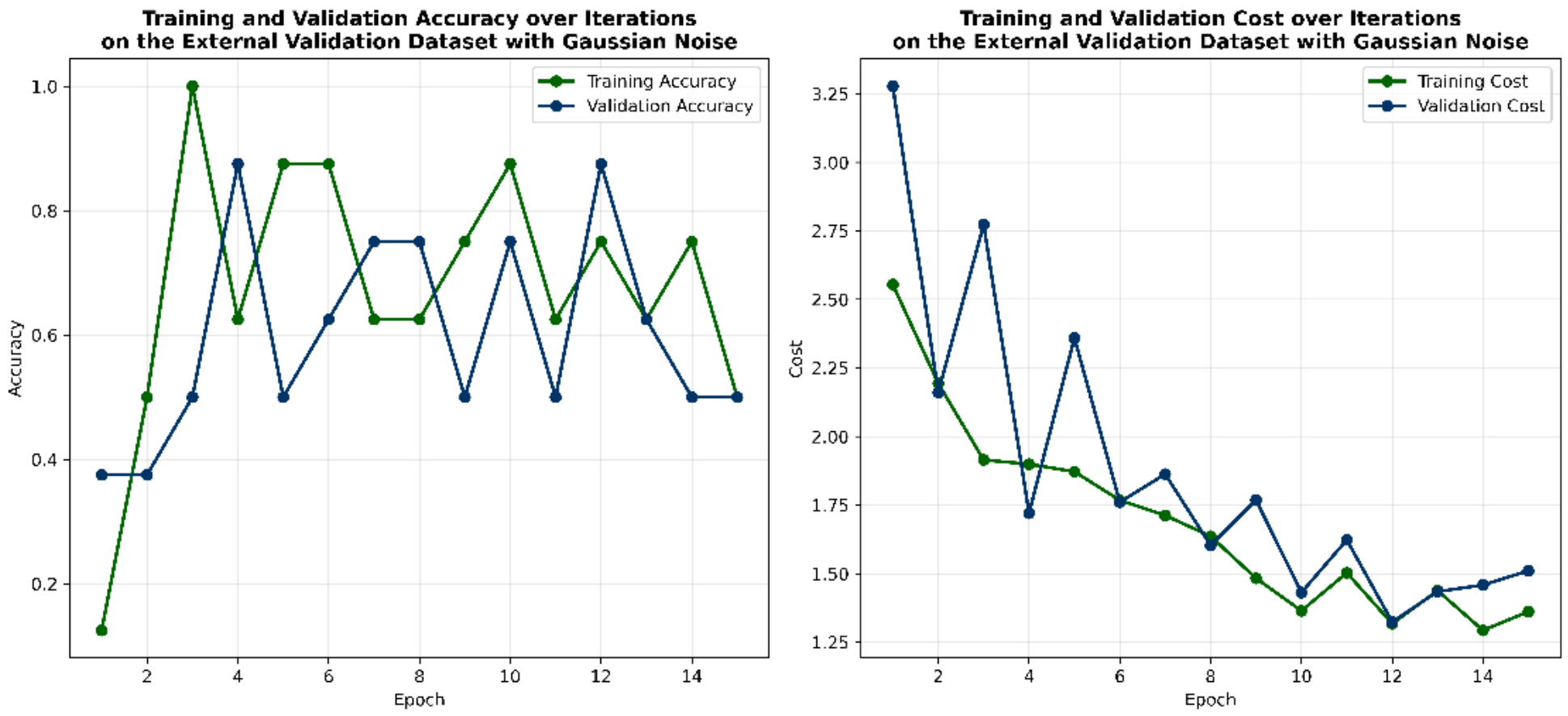

**Figure 13.** Visualization of the training and validation accuracy and loss curves obtained with the proposed QBPM following the addition of Gaussian noise to the MR images in the external validation dataset.

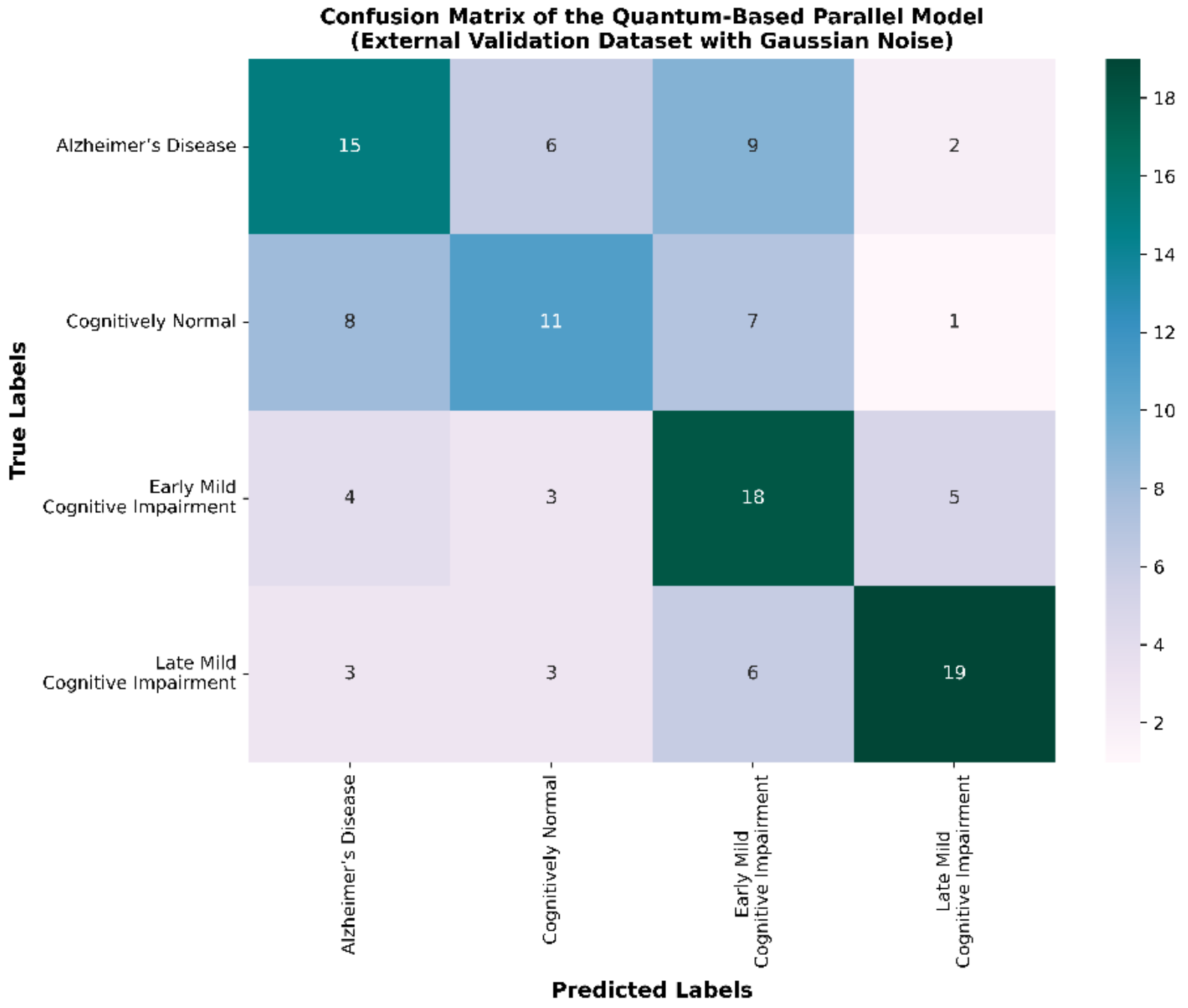

**Figure 14.** Confusion matrix illustrating the classification performance of the model on the external validation dataset after applying Gaussian noise.

For the four stages, the precision, recall, and $F_1$-score range from 0.45 to 0.70, 0.41 to 0.61, and 0.44 to 0.66, respectively, with detailed results provided in **Table 3**.

### *3.6. Results obtained when artificial gate-level noise is introduced to the quantum circuit on the external validation dataset*

In the previous section, gate-level artificial noise was added to the QBPM circuit for the original dataset. Here, the same procedure was applied to the external validation dataset to analyze the model's robustness to noise. Similarly, Gaussian noise was introduced into the proposed quantum circuit, and the model's classification performance and noise resilience were evaluated under these conditions. Upon executing the model, the average training accuracy, validation accuracy, training loss, and validation loss on the external validation dataset were 0.81, 0.70, 2.45, and 2.11, respectively. The corresponding accuracy and loss curves are presented in **Figure 15**, while the associated confusion matrix is shown in **Figure 16**.

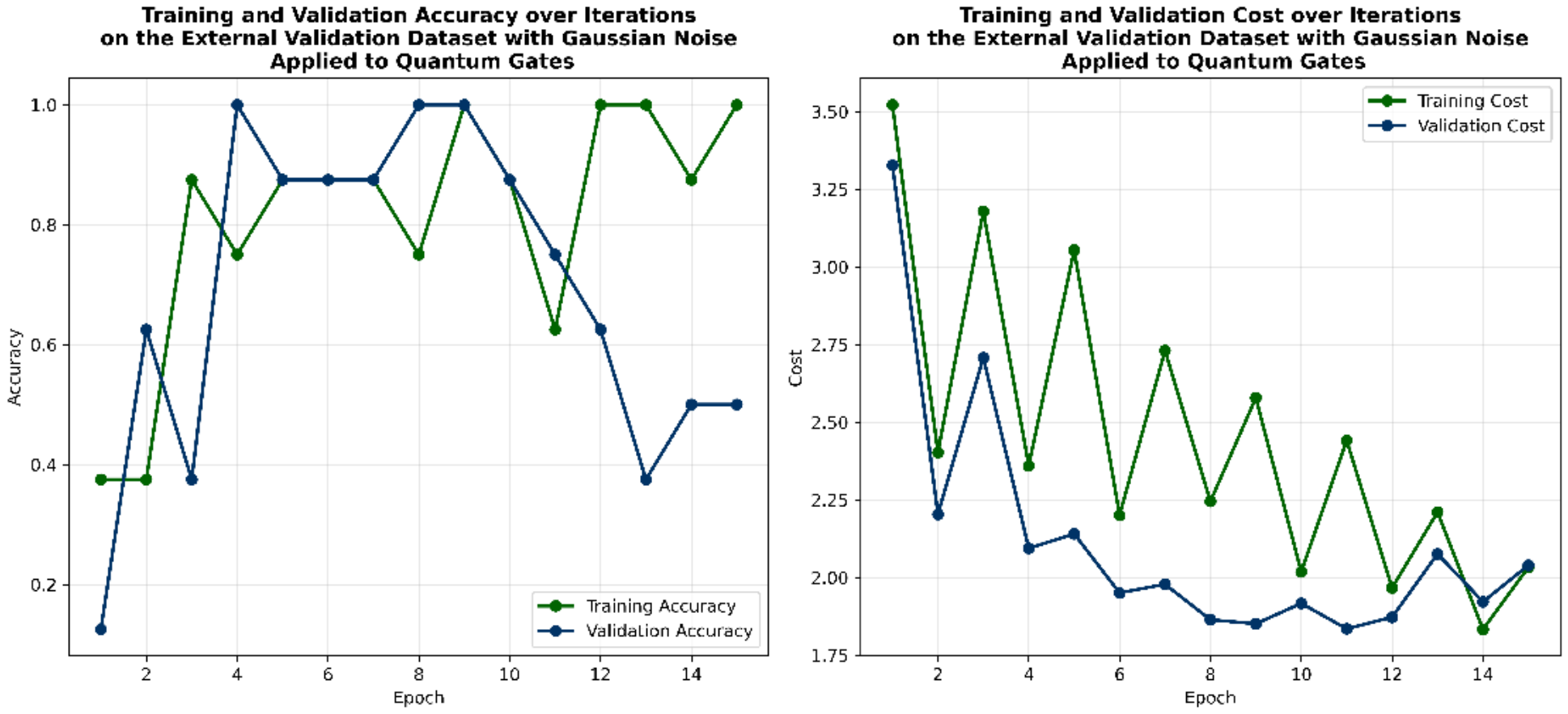

**Figure 15.** Training and validation accuracy and loss curves on the external validation dataset obtained with the proposed QBPM following the addition of artificial noise to the quantum gates.

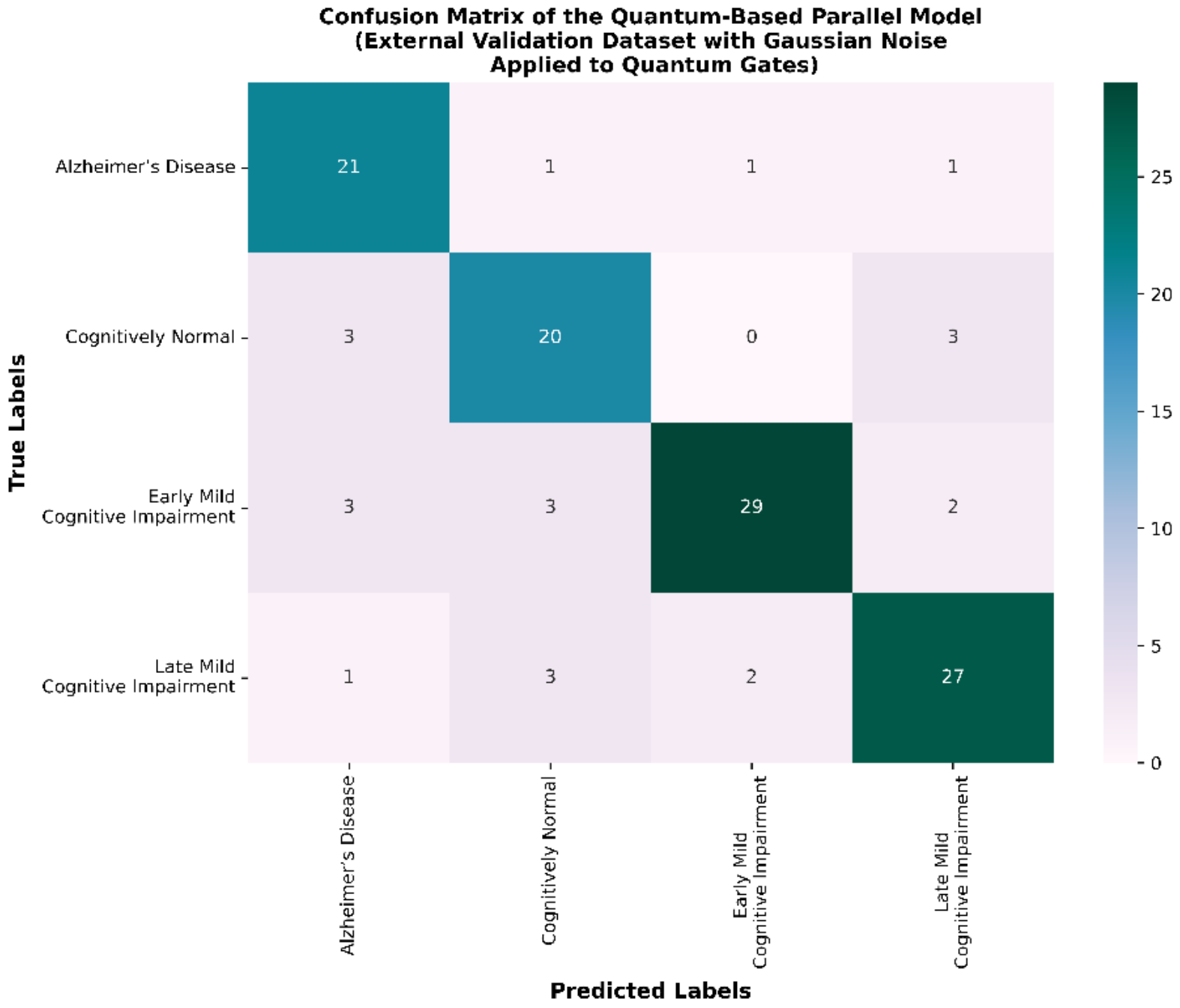

**Figure 16.** Confusion matrix obtained with the proposed QBPM following the addition of artificial noise to the quantum gates.

For the four classes, the precision, recall, and $F_1$-score range from 0.74 to 0.91, 0.77 to 0.88, and 0.75 to 0.84, respectively, with detailed results provided in **Table 3**.

**Table 2.** Detailed metrics of the classification performance of the proposed QBPM on the original dataset.

| | Stages of AD | Precision | Recall | $F_1$-Score |
|---|---|---|---|---|
| **QBPM architecture-Original Dataset** | Very Mild Dementia | 0.91 | 0.78 | 0.84 |
| | Mild Dementia | 0.81 | 0.81 | 0.82 |
| | Moderate Dementia | 0.84 | 0.98 | 0.90 |
| **QBPM architecture-Gaussian Noise on Original Dataset** | Very Mild Dementia | 0.80 | 0.93 | 0.86 |
| | Mild Dementia | 0.93 | 0.52 | 0.67 |
| | Moderate Dementia | 0.88 | 0.94 | 0.91 |
| **QBPM architecture-Gate-Level Noise on Quantum Circuit** | Very Mild Dementia | 0.83 | 0.81 | 0.82 |
| | Mild Dementia | 0.73 | 0.73 | 0.74 |
| | Moderate Dementia | 0.84 | 0.86 | 0.85 |

**Table 3.** Detailed metrics of the classification performance of the proposed QBPM on the external validation dataset.

| | Stages of AD | Precision | Recall | $F_1$-Score |
|---|---|---|---|---|
| **QBPM architecture-External Validation Dataset** | Cognitively Normal | 0.84 | 0.68 | 0.75 |
| | Early Mild Cognitive Impairment | 0.74 | 0.81 | 0.78 |
| | Late Mild Cognitive Impairment | 0.81 | 0.69 | 0.75 |
| | Alzheimer's Disease | 0.67 | 0.88 | 0.76 |
| **QBPM architecture-Gaussian Noise on External Validation Dataset** | Cognitively Normal | 0.48 | 0.41 | 0.44 |
| | Early Mild Cognitive Impairment | 0.45 | 0.60 | 0.51 |
| | Late Mild Cognitive Impairment | 0.70 | 0.61 | 0.66 |
| | Alzheimer's Disease | 0.50 | 0.47 | 0.48 |
| **QBPM architecture-Gate-Level Noise on Quantum Circuit** | Cognitively Normal | 0.74 | 0.77 | 0.75 |
| | Early Mild Cognitive Impairment | 0.91 | 0.78 | 0.84 |
| | Late Mild Cognitive Impairment | 0.82 | 0.82 | 0.83 |
| | Alzheimer's Disease | 0.75 | 0.88 | 0.81 |

### *3.7. Results of classical baseline models for comparison*

In this section of the study, five classical TL methods—EfficientNetB0, InceptionV3, MobileNetV2, ResNet50, and VGG16—were employed to compare the classification performance of the proposed QBPM architecture in distinguishing AD classes. The performance of these models was evaluated in terms of training loss, validation loss, training accuracy, and validation accuracy. The variations across iterations are illustrated in **Figure 17**, while the confusion matrices visualizing the classification performance of the models are presented in **Figure 18**.

For the EfficientNetB0 model, the average training accuracy and validation accuracy were 0.85 and 0.86, respectively, with training and validation losses of 0.37 and 0.32. Precision, recall, and $F_1$-score values for the classes Mild Dementia, Moderate Dementia, and Very Mild Dementia ranged from 0.84–1.00, 0.57–1.00, and 0.71–0.99, respectively (**Table 4**). The InceptionV3 model achieved an average training accuracy of 0.78 and validation accuracy of 0.86, with corresponding training and validation losses of 0.87 and 0.37. For the three classes, precision, recall, and $F_1$-score ranged from 0.76–1.00, 0.39–1.00, and 0.56–0.99, respectively (**Table 4**). The MobileNetV2 model demonstrated strong performance, achieving training and validation accuracies of 0.85 and 0.86, with training and validation losses of 0.35 and 0.31. Precision, recall, and $F_1$-score values ranged from 0.88–1.00, 0.64–1.00, and 0.78–1.00, respectively (**Table 4**). ResNet50 achieved the highest classification accuracy, with average training and validation accuracies of 0.87 and 0.90, and training and validation losses of 0.30 and 0.26. For this model, precision, recall, and $F_1$-score ranged from 0.89–1.00, 0.63–1.00, and 0.75–1.00, respectively (**Table 4**). Finally, the VGG16 model yielded comparable results, with training and validation accuracies of 0.79 and 0.86, and training and validation losses of 0.58 and 0.35. Its precision, recall, and $F_1$-score values ranged from 0.71–1.00, 0.31–1.00, and 0.47–0.99, respectively (**Table 4**).

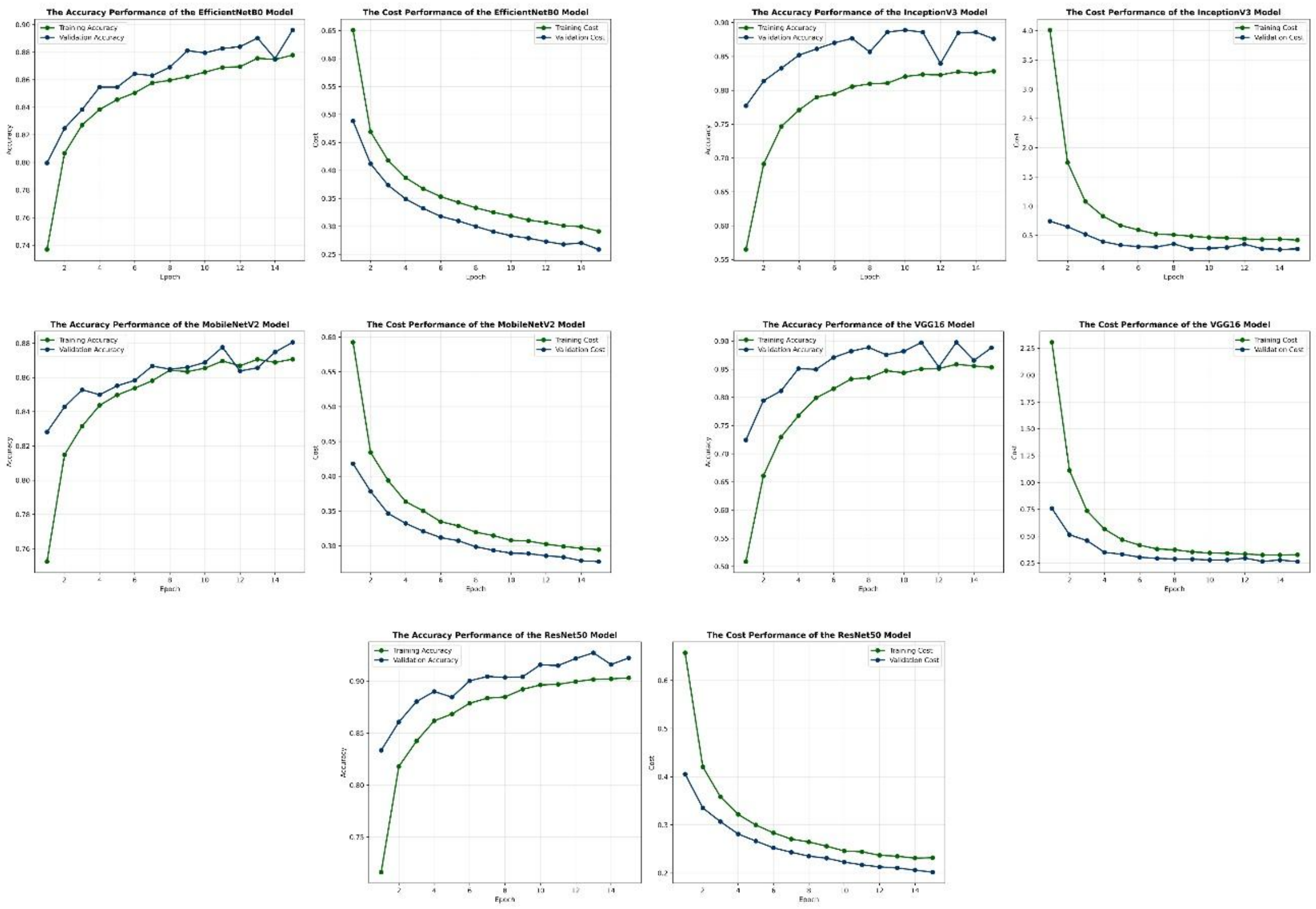

**Figure 17.** Epoch-wise changes in training and validation accuracy and loss values of five classical TL models, compared with the proposed QBPM architecture, for the classification of AD.

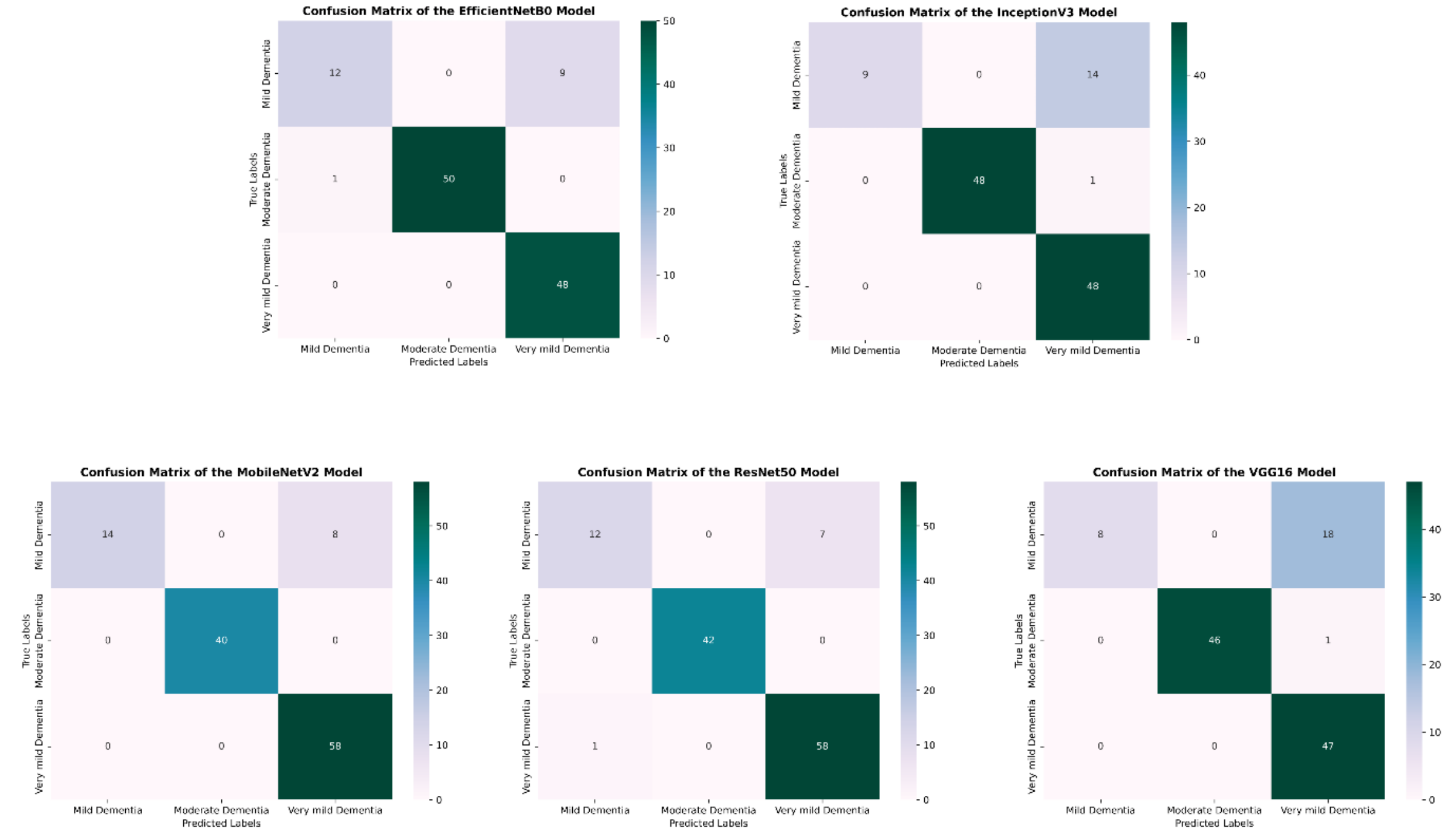

**Figure 18.** Confusion matrices of EfficientNetB0, InceptionV3, MobileNetV2, ResNet50, and VGG16 models obtained for the classification of AD.

**Table 4.** Class-wise precision, recall, and $F_1$-score values of EfficientNetB0, InceptionV3, MobileNetV2, ResNet50, and VGG16 models in the classification of AD.

| | **Stages of AD** | **Precision** | **Recall** | **$F_1$-Score** |
|---|---|---|---|---|
| **EfficientNetB0** | Very Mild Dementia | 0.84 | 1.0 | 0.91 |
| | Mild Dementia | 0.92 | 0.57 | 0.71 |
| | Moderate Dementia | 1.0 | 0.98 | 0.99 |
| **InceptionV3** | Very Mild Dementia | 0.76 | 1.0 | 0.86 |
| | Mild Dementia | 1.0 | 0.39 | 0.86 |
| | Moderate Dementia | 1.0 | 0.98 | 0.99 |
| **MobileNetV2** | Very Mild Dementia | 0.88 | 1.0 | 0.94 |
| | Mild Dementia | 1.0 | 0.64 | 0.78 |
| | Moderate Dementia | 1.0 | 1.0 | 1.0 |
| **ResNet50** | Very Mild Dementia | 0.89 | 0.98 | 0.94 |
| | Mild Dementia | 0.92 | 0.63 | 0.75 |
| | Moderate Dementia | 1.0 | 1.0 | 1.0 |
| **VGG16** | Very Mild Dementia | 0.71 | 1.0 | 0.83 |
| | Mild Dementia | 1.0 | 0.31 | 0.47 |
| | Moderate Dementia | 1.0 | 0.98 | 0.99 |

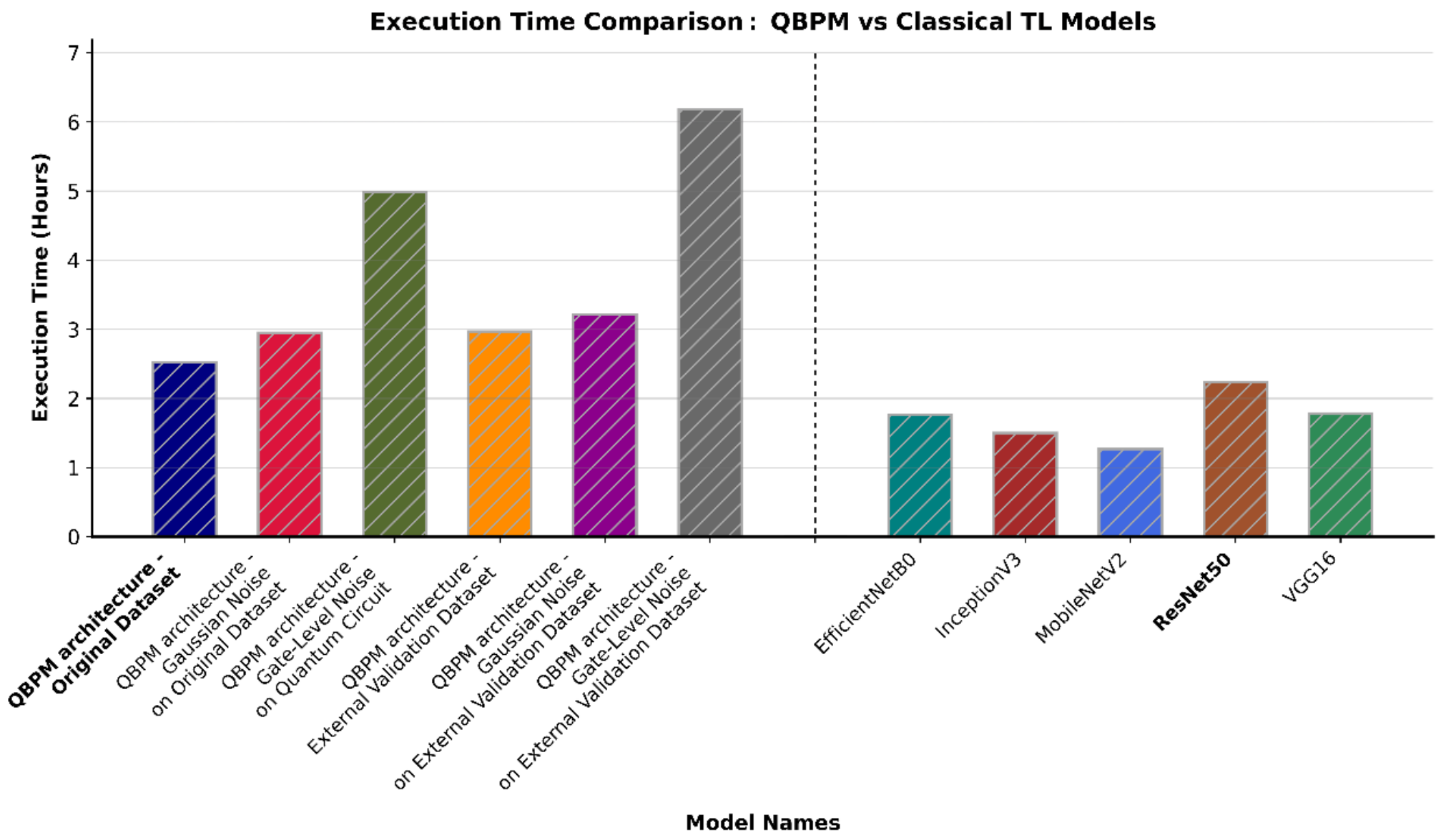

**Figure 19.** Comparison of execution times between QBPM variants and classical TL models. The bar chart illustrates the total execution times (in hours) for various QBPM architecture configurations—including the original dataset, the external validation dataset, and both datasets subjected to Gaussian noise, as well as gate-level noise applied to the quantum circuits—and compares them with widely used 5 different classical TL models (EfficientNetB0, InceptionV3, MobileNetV2, ResNet50, and VGG16). As shown, there is no significant difference in total execution time between the QBPM architecture run on the original dataset and ResNet50, which achieves the highest classification performance among the classical TL models.

## 4. Discussion

With the advancement of modern medicine and the consequent extension of human lifespan, AD has emerged as one of the most significant neurodegenerative threats, attracting substantial attention from both the scientific community and the general public. Given that AD leads to severe cognitive and behavioral impairments in later stages, early-stage automated diagnosis and detection, coupled with effective therapeutic interventions, can slow disease progression and significantly enhance patients' quality of life. Currently, AI-based decision support systems are widely employed for early diagnosis and staging of AD, particularly utilizing neuroimaging techniques such as MRI, achieving notable levels of success. For instance, in the study by [33], a DL model named DEMentia NETwork (DEMNET) was proposed, designed from scratch with fewer model parameters, making it suitable for training on small datasets. The model was applied to MR images obtained from the Kaggle dataset, encompassing classes of Very Mild Dementia (VMD), Mild Dementia (MID), Moderate Dementia (MOD), and Non-Demented (ND). The proposed architecture consisted of convolutional layers, max-pooling, dropout, dense layers, and four DEMNET blocks. Each DEMNET block comprised two convolutional layers, a batch normalization layer, and a max-pooling layer. By addressing the class imbalance problem using the SMOTE technique, the model achieved 95.23% accuracy, 97% AUC, and a Cohen's Kappa of 0.93. In another study [38], pre-trained architectures including ResNet101, Xception, and InceptionV3 on the ImageNet dataset were employed for automatic detection and classification of AD from brain MR images. These models were evaluated using two strategies: freezing the initial blocks and retraining the remaining layers through transfer learning, and training all layers from scratch. MR images were segmented into White Matter (WM), Gray Matter (GM), and Cerebrospinal Fluid (CSF) regions using a U-Net architecture, allowing a comparison of classification performance between models trained on both segmented and whole MRI images. Results demonstrated that, across both internal (ADNI) and independent test sets (OASIS and AIBL), the InceptionV3-based TL model achieved the highest accuracy when whole MR images were used as input. Moreover, CNN models utilizing TL exhibited superior classification performance compared to models trained from scratch on limited data. For multi-class classification of AD based on MR images, a customized Convolutional Neural Network (CNN) was proposed in [34]. During preprocessing, T1-weighted MR images from the ADNI database were processed using Statistical Parametric Mapping 12 (SPM12) and segmented into GM, WM, and CSF. Classification was subsequently performed using only GM slices. The proposed model comprised convolutional layers, four dense blocks, transition layers, and a fully connected classifier. By keeping the early layers fixed and retraining the last two blocks, the model successfully classified multiple stages of AD with 97.84% accuracy.

According to the literature review conducted in this study, direct performance comparisons between studies are generally not feasible due to variations in datasets used for AD classification, imaging modalities (e.g., structural MRI, PET, genomic data), preprocessing steps that may influence classification outcomes, model hyperparameters, and task definitions (e.g., binary versus multi-class classification). Therefore, in the present work, five classical TL models—EfficientNetB0, InceptionV3, MobileNetV2, ResNet50, and VGG16—were comparatively evaluated under identical experimental conditions (e.g., fixed number of epochs, batch size, and regularization parameters) with the proposed QBPM architecture. Among the TL models compared, ResNet50 achieved the highest classification accuracy and the lowest loss in distinguishing the very mild, mild, and moderate dementia stages of AD, demonstrating the most successful performance.

In this study, QBPM architecture is proposed, inspired by the principle of classical model parallelism, to achieve high-accuracy discrimination of AD stages based on MRI scans. In the proposed architecture, two distinct PQC models are executed in parallel on two quantum simulators, with the aim of capturing the complex patterns and pixel-level high-dimensional information within MRI data, as well as structural alterations that play a critical role in differentiating AD stages, such as brain atrophy, gray matter loss, and ventricular enlargement. An important point to emphasize is that, although both parallel PQC models are trained on the same sample batch, parameter optimization processes are conducted independently owing to parallel simulation. This design enables exploration of diverse parameter spaces, thereby substantially enhancing the learning capacity. Subsequently, the measurement outcomes obtained from the two parallel PQC models are integrated to improve classification performance. In both PQC models, the computation begins with the application of $Hadamard$ gates to the input data expressed as quantum states, allowing richer representation and processing of information. In the first PQC model, over 20 layers, parameterized quantum gates including $U3$, $R_x$ and $R_y$, are applied, followed sequentially by fixed non-parameterized gates, namely $CNOT$ and $CCNOT$. In this process, the intricate and pixel-level information embedded in MR images represented as quantum states is effectively learned through the parameterized $U3$, $R_x$ and $R_y$ gates. To reduce the number of trainable parameters and accelerate the optimization process, a "parameter-sharing strategy" is employed for the $R_x$ and $R_y$ rotational gates, ensuring shared parameters across the circuit. Furthermore, the strong correlation information between pixels is modeled via different entanglement topologies constructed with $CNOT$ and $CCNOT$ gates and preserved through transmission across entangled qubits. Using the two-qubit $CNOT$ gate, both "dense all-to-all entanglement" and "cyclically repeated nearest-neighbor entanglement" are implemented, establishing both local and long-range interactions among qubits. This allows spatial correlation information in the data to be more effectively retained within the quantum circuit. Similarly, the second PQC model follows the same parameterization over 20 layers; however, unlike the first model, the correlation information across all qubits is captured using the two-qubit $CY$ gate. Through these operations, a stronger entanglement structure is established in the circuit, further enhancing the capacity to represent inter-qubit dependencies.

In the proposed QBPM architecture, the two parallel PQC models executed simultaneously on the same quantum simulator result in a total of 1,803 trainable parameters, calculated as *num_layers × num_qubits × trainable parameters × 2 + bias = 20 × 15 × 3 × 2 + 3 = 1803*. Among the classical TL models, the pre-trained ResNet50 architecture achieved the highest classification accuracy (average training accuracy: 0.87, validation accuracy: 0.90) and the lowest loss values (training loss: 0.30, validation loss: 0.26). While the original ResNet50 architecture comprises approximately 23.5 million trainable parameters, in this study all layers except the final layers were frozen to shorten training time and mitigate overfitting, reducing the number of trainable parameters to approximately 6,000. As a result, in the classification of AD, the proposed QBPM architecture demonstrated a remarkable achievement by attaining ~3% higher training and validation accuracy compared to ResNet50, while requiring nearly four times fewer trainable parameters.

Although the quantum model exhibited higher loss values than its classical counterpart due to the optimization processes involved, the substantially reduced number of trainable parameters alongside superior accuracy highlights the proposed architecture as a strong alternative in terms of classification performance. Furthermore, several studies in the literature emphasize that quantum models achieve higher performance than classical models despite utilizing significantly fewer parameters [106–109]. In terms of total execution time (**Figure 19**), there was no significant difference between the ResNet50 model (2 hours 14 minutes) and the proposed model (2 hours 31 minutes), although ResNet50 demonstrated slightly better efficiency in runtime.

An additional ablation study was conducted to experimentally substantiate the advantages of the parallel quantum design in the proposed QBPM model and to validate the necessity of the parallel architectural design. In this ablation study, Circuit 1 (only the first circuit in the parallel architecture), Circuit 2 (only the second circuit in the parallel architecture), the Sequential Model (the two circuits in the parallel architecture applied sequentially), and the proposed QBPM model were evaluated on the original dataset. In order to completely eliminate all sources of uncertainty in the modeling and training processes and to ensure that all models could be compared in a fair, unbiased, and fully reproducible experimental environment, parameter initializations, dataset splitting, shuffling, and mini-batch selection processes were each independently fixed using the SEED=42 parameter. Thus, it was ensured that all components of the simulation environment operated in a fully reproducible manner. As a result of the experimental evaluations, the effective circuit depth (the maximum depth executed sequentially per quantum channel), total number of gates, total execution times (simulator wall-clock time), and the mean training accuracy, validation accuracy, training loss, and validation loss calculated over 15 training/validation iterations were obtained for the four different model architectures. All of these metrics are presented in detail in **Supplementary Table 1**. In addition, the corresponding accuracy and loss graphs for each iteration of Circuit 1, Circuit 2, the Sequential Model, and the proposed QBPM model are presented in **Supplementary Figures 1–4**, respectively, while the corresponding confusion matrices visualizing the classification performance are provided in **Supplementary Figures 5–8**, respectively. Furthermore, the precision, recall, and $F_1$-score values achieved by the models for the three stages of Alzheimer's disease are presented in **Supplementary Table 2**.

When the obtained results were examined, it was determined that the total execution times of the models corresponding to the individual Circuit 1 and Circuit 2 structures were lower than that of the proposed QBPM architecture. However, due to their limited feature representation capacities, these individual models were able to achieve lower mean validation accuracies. The Sequential model, on the other hand, exhibited the highest execution time and the lowest validation accuracy compared with all other models. Connecting both circuit structures sequentially doubles the effective circuit depth, which may adversely affect gradient optimization, extend the parameter update process, and consequently degrade validation performance. In contrast, the proposed parallel model approach achieved a superior validation accuracy compared with all other alternatives, demonstrating that it successfully enhanced quantum feature representation without being exposed to the trainability penalties introduced by the Sequential model. Through its parallel structure, the proposed model significantly increased its learning capacity by simultaneously exploring different parameter spaces. By keeping the effective circuit depth low, it enhanced both trainability and circuit expressibility while completing the training process within an acceptable execution time (**Supplementary Figures 9**).

Analysis of the convergence curves reveals that the validation accuracy of the proposed QBPM model increases with notable stability, reaching 1.00 accuracy as early as the 6th iteration, thereby exhibiting the fastest convergence among the evaluated architectures. The model reached an accuracy of 1.00 from the 8th iteration onward, forming a plateau and maintaining this stability as the training process progressed. In contrast, the individual Circuit 1 and Circuit 2 models exhibited sharp performance decreases, particularly at the 4th and 7th iterations, resulting in a fluctuating convergence profile. In the Sequential model, severe initial stagnation, which was not observed in the QBPM architecture, was also observed. Consequently, when the iteration-based variations in validation accuracy are evaluated in general, it can be stated that the QBPM architecture is the model that converges the fastest, exhibits the most stable progression, and achieves the highest accuracy values. Similarly, when the convergence of the validation loss curves was examined, it was determined that the QBPM architecture settled into a lower and narrower oscillation band from the 8th iteration onward.

A critical point that should be emphasized here is the difference between the results initially obtained by the proposed QBPM architecture on the original dataset (**Table 2**) and the accuracy and execution time values obtained by the QBPM architecture in the ablation study conducted under a fixed reproducible seed environment (**Supplementary Table 1**). The performance metrics and execution times of PQC-based models are directly dependent on the position of the initial weights in the parameter space and the optimization convergence path. In the ablation study, since parameter initializations, dataset splitting, shuffling, and mini-batch selection processes were all fixed using SEED=42, the model produces fully reproducible results in the parameter and optimization spaces regardless of how many times the simulation is executed. The use of a fixed seed ensured consistent initialization and batch ordering, thereby promoting more stable and reproducible convergence behavior and contributing to more consistent execution times. In contrast, in the initial experimental scenario, since the processes, including parameter initialization, were performed randomly, the model may have become trapped in local minima within the parameter space; this may have delayed optimization convergence and consequently extended the total execution time. Furthermore, under random initialization, the model's tendency to focus on noisy patterns in the data rather than general characteristics during the early stages of training may have led to small fluctuations in generalization performance. Therefore, these behavioral differences between the two experimental environments are fully consistent with and expected from the nature of parameterized quantum algorithms.

Furthermore, to empirically evaluate the barren plateau dynamics and theoretical scalability of the proposed QBPM architecture as the number of qubits increases, the variance of the cost function gradient was analyzed across increasing qubit counts. In accordance with the theoretical framework established in Equations (24) and (25), the scaling behavior of the cost function gradient variance was investigated to assess the optimization landscape for potential barren plateau formation. The resulting gradient variance profiles before and after training for qubit counts of $n = 10, 12,$ and $15$ are presented in **Supplementary Figure 10**. Prior to training (random initialization), the cost function gradient variance remains above the defined threshold and exhibits a slight upward trend as the qubit count increases. Following training (with optimized parameters), the gradient variance increases more pronouncedly with increasing qubit capacity. This behavior indicates that the optimization process preserves, and even enhances, the magnitude of the gradients across the investigated qubit configurations, thereby mitigating the risk of vanishing gradients and supporting effective parameter updates. Importantly, at both stages, the model consistently remains within the designated trainable region, with no evidence of the exponential gradient-variance decay toward zero that characterizes barren plateau formation. Consequently, the proposed QBPM architecture maintains trainability and optimization stability across both the initial and trained states within the investigated qubit range, providing empirical evidence of its scalability and robustness against barren plateau formation.

In addition, to evaluate the applicability of the proposed QBPM architecture under real-world conditions and to simulate the noise effects originating from quantum gates in actual quantum computing environments, the model was tested under two distinct noise scenarios. In the first scenario, Gaussian or normally distributed noise was added to the MR images in order to mimic artifacts that may arise from patient head movements during scanning, operator-related errors, or imaging device imperfections. Specifically, random values drawn from a Gaussian distribution with mean 0 and standard deviation 0.01, scaled by a noise factor of 0.5, were added to each pixel of the MR images. Importantly, the relatively high noise factor was deliberately chosen to rigorously assess the model's reliability and robustness under realistic conditions. Under this setting, the model exhibited a moderate decline in training accuracy (90% → 87%) and validation accuracy (93% → 88%), accompanied by a slight increase in training loss (1.65 → 1.68) and validation loss (1.85 → 1.92). Nevertheless, these changes remained within expected ranges and at tolerable levels. Notably, despite the severe distortions in MR images caused by the high-intensity noise, the model's learning performance was largely preserved. In the second scenario, Gaussian noise was applied to the quantum gates within the simulation environment to partially reflect the gate-level errors typically encountered in real quantum hardware and to examine their impact on classification performance. In this case, random values drawn from a Gaussian distribution with mean 0 and standard deviation 0.01 were added to the angle parameters of the parameterized quantum gates ($U3$, $R_x$ and $R_y$). Furthermore, since it is not feasible to directly inject noise into multi-qubit gates ($CNOT$, $CY$ and $CCNOT$), which are responsible for carrying correlation information across pixels, these gates were reformulated using the $R_z$ rotation gate, thereby incorporating phase noise—one of the most prevalent types of quantum noise—into the model. Under these conditions, the model's training accuracy dropped from 90% to 83% and validation accuracy from 93% to 77%, while training loss increased from 1.65 to 1.83 and validation loss from 1.85 to 2.28. Although these results are lower compared to the accuracy and loss values obtained when Gaussian noise is added to the MR images, the fact that the decline remains at a moderate level and that the proposed model continues to outperform many classical models even in noisy environments demonstrates and experimentally validates the robustness of the model.

When examining the training and validation accuracy and loss curves, clear changes in the model's learning behavior were observed as the scenarios progressed from the normal condition to those in which Gaussian noise was sequentially added to the images and quantum gates. Specifically, when Gaussian noise was applied to the quantum gates, both training and validation accuracies displayed more fluctuation and a less stabilized trajectory compared to the normal condition. In contrast, under the normal scenario, validation accuracy began to decline from around the 12th epoch, indicating a risk of "late-stage overfitting". When noise was introduced to the quantum gates, this overfitting risk appeared reduced, with the model exhibiting a more balanced and generalizable learning tendency. In other words, despite the presence of noise, the model maintained a more stable learning process and avoided overfitting. The overfitting observed under normal conditions is likely attributable to the limited size of the training dataset. Although the proposed QBPM architecture is highly complex, the relatively small size of the dataset used for training and validation constrains the model's generalization capacity. Furthermore, in addition to dataset size, the presence of class imbalance constitutes another significant factor that could negatively affect the model's learning performance. Based on these findings, it can be concluded that introducing a certain level of Gaussian noise to the quantum gates both reduces overfitting risk and enhances the model's generalization ability. On the other hand, examining the loss curves reveals that the gap between training and validation losses increases as the scenario transitions from the normal condition to the case with noise applied to the quantum gates. While noise reduces overfitting risk, it induces a certain degree of instability in the model outputs. By optimizing the noise factor, this risk can be mitigated to a controllable extent.

In this study, the proposed QBPM architecture was further evaluated on an additional external validation dataset, namely ADNI: Alzheimer's MRI Classification, to assess its overall robustness, generalization capability, and performance against MRI data exhibiting diverse demographic and clinical characteristics. Within this framework, the model was employed to distinguish among the classes of Cognitively Normal, Early Mild Cognitive Impairment, Late Mild Cognitive Impairment, and Alzheimer's Disease. Under these conditions, the model demonstrated superior classification performance compared to many conventional models, achieving accuracies of 0.85 and 0.81 and loss values of 1.66 and 1.86 on the training and validation datasets, respectively. These results indicate that the proposed quantum-based model successfully learns from MR images representing diverse patient profiles, exhibiting high performance in terms of both overall robustness and generalization. Moreover, the absence of the late-stage overfitting trend observed in the original dataset within the external validation dataset, together with the more stable learning trajectory in the training/validation curves and the consistent structure observed in the training/validation cost plots, supports the conclusion that the model operates in a coherent and reliable manner across different data cohorts. However, one important point to be emphasized here is that the external dataset has a more class-balanced structure compared to the original dataset. Under conditions where Gaussian noise is introduced to the quantum gates, the model exhibits markedly increased instability and a substantial decline in generalization capability. Additionally, in both the original and external validation datasets, the introduction of Gaussian noise to the quantum gates significantly prolongs execution times (Figure 20), indicating that the model requires more computational resources during the optimization process. These observations underscore the considerable impact of noise on model performance and highlight the necessity of carefully choosing the noise factor to maintain optimal outcomes.

In the literature, a multitude of quantum AI-based approaches have been proposed for the classification of AD using similar datasets, achieving varying degrees of success. For instance, in the study by [71], a deep ensemble learning model integrated with a quantum ML classifier was proposed for the classification of AD based on MR images. Preprocessing steps, including cropping, resizing, denoising, and data augmentation, were applied to images obtained from the ADNI1 and ADNI2 datasets. Following dataset integration, a fine-tuned and customized deep ensemble approach combining VGG16 and ResNet50 models was employed for feature extraction. Features extracted from these two models were concatenated and fed into a Quantum Support Vector Machine (QSVM) algorithm implemented using the Qiskit library, enabling successful classification of AD into four distinct stages. The results demonstrated that the proposed VGGNet + ResNet + QSVM approach outperformed both individual DL models (VGGNet, ResNet) and classical SVM classifiers integrated with these models, achieving 99.89% accuracy and a 99.99% AUC. In another study [67], a hybrid model based on classical DL and quantum ML was proposed for AD classification using ADNI MR images, as well as for comparative analysis on a brain tumor dataset. The classical component of the model consisted of three convolutional blocks followed by a fully connected layer, after which a hybrid quantum layer transformed classical data into quantum states, facilitating speed and efficiency in the feature extraction process. Experimental results indicated that the proposed model achieved higher accuracy in both Alzheimer's and brain tumor classification compared to classical CNN architectures. Another investigation [70] proposed a hybrid classical–quantum deep neural network for AD detection based on MR images, capable of distinguishing between demented and non-demented individuals. In the classical component, the ResNet34 architecture was employed to extract a 512-dimensional feature vector from the input images. This vector was then encoded into a variational quantum circuit, where data embedding was performed using $Hadamard$, $R_y$, and $R_z$ gates, entanglement was established via $CNOT$ gates, and measurements were conducted. The vector, reduced to dimension four within the quantum circuit, was subsequently classified into two classes via a fully connected layer. The hybrid model was executed on PennyLane (default.qubit), Qiskit Aer, and Qiskit BasicAer simulators; comparative analysis revealed that the PennyLane default simulator achieved 97% test accuracy, and the model outperformed purely classical architectures such as GoogLeNet and ResNet34. Furthermore, in this study [110], the Modified Variational Quantum Classifier (MVQC) model, developed for the early detection of AD, was introduced. Unlike standard VQC architectures, shallow parameterized circuits were employed in conjunction with a customized feature mapping method incorporating denoising, PCA-based dimensionality reduction, and amplitude encoding. The data dimension was reduced to four qubits, and the model was executed on the IBM QASM simulator. Shallow parameterized circuits and reduced entanglement layers were utilized to mitigate noise and error propagation, thereby enhancing stability. Moreover, the model's generalization capacity was assessed using five-fold cross-validation. In this study, where stage I and stage II Alzheimer's neuroimaging datasets were combined, the MVQC model achieved 93% accuracy with 40% shorter training time compared to classical CNNs, and under five-fold cross-validation, it attained 99.12% AUC and 93.45% $F_1$-score.

In conclusion, the proposed QBPM architecture demonstrated superior performance in classifying stages of AD compared to numerous classical TL methods, despite employing a relatively low number of trainable parameters. Its comparable total execution times with classical approaches further position the model as a time-efficient alternative. The model's robust performance on an external validation dataset comprising diverse patient profiles underscores its resilience and high generalization capacity. Furthermore, gradient variance analysis confirmed that the model remains free from the barren plateau problem, theoretically guaranteeing trainability and scalability as qubit counts increase. In particular, even in scenarios where noise effects arising from both real-world conditions and quantum gates in the quantum computing environment are simulated with high-level Gaussian noise, the model has demonstrated superior performance compared to many classical methods. These findings indicate that the proposed model is not only theoretically robust but also a strong candidate for practical applications, offering effective utility in the early detection of complex diseases such as Alzheimer's.

## 5. Conclusions, Limitations and Future Works

To the best of our knowledge, this study provides the first comprehensive investigation of a QBPM architecture, developed by drawing inspiration from the principle of classical model parallelism, for classifying AD stages. The performance of the proposed model was evaluated on both the original and external validation datasets. The proposed model demonstrated high classification performance across both datasets, highlighting its overall robustness and strong learning capacity, while its performance on MRI data from patient profiles with diverse demographic and clinical characteristics provided experimental evidence of its generalization capability. Additionally, to evaluate the scalability and trainability of the proposed QBPM architecture, the scaling behavior of the cost function gradient variance was analyzed before and after training; both empirical and theoretical results demonstrated that the architecture operates completely free from barren plateau risks within the evaluated qubit range. Furthermore, when the proposed model was run in environments in which both real-world conditions and gate-induced errors occurring in real quantum computing systems were simulated with high levels of Gaussian noise, its classification performance exhibited a certain decline in both noise scenarios; however, this decline remained moderate, thereby demonstrating the model's applicability not only in theory but also under practical conditions. Finally, when compared with five different classical TL methods, the proposed QBPM architecture experimentally demonstrated superior performance with substantially fewer circuit parameters, establishing it as a powerful alternative to classical models in terms of both classification accuracy and execution time.

In the conclusion section of the manuscript, it is appropriate to acknowledge certain potential limitations of the study as follows:

- **Error correction codes:** Since the proposed QBPM architecture was executed using the default.qubit state vector simulator provided by the PennyLane framework, error correction codes were not employed in this study. Consequently, the implementation of error correction codes falls beyond the scope of the present work. In order to analyze the reliability of the proposed model and its behavior under realistic conditions in greater detail, future studies aim to use actual quantum computers, thereby enabling the integration of error correction codes.
- **Dataset limitation:** The classification performance of the proposed model in this study was evaluated solely on two different AD datasets. To more comprehensively assess the model's robustness and generalization capability in real-world applications, further testing on larger datasets obtained from diverse sources, encompassing a variety of demographic and clinical characteristics, is required.

In future studies, it is intended to transform the proposed Quantum-Based Parallel Model architecture into an "end-to-end quantum artificial intelligence system" that can integrate and make decisions based not only on MR images but also on multimodal information such as genomic, clinical, and PET data from the same patients. Additionally, by running the model in a real quantum computing environment, we aim to demonstrate its applicability and robustness under real-world conditions, beyond its theoretical success.

## Authors' contributions

All authors contributed to the study conception and design. Material preparation, data collection, software implementation, validation, formal analysis, investigation, resources, data curation, visualization, and writing—original draft preparation were performed by EA. Writing—review and editing was carried out by EA and MO. Funding acquisition was provided by EA and MO. Supervision was carried out by MO. All authors read and approved the final manuscript.

## Funding

This study was supported by the Scientific and Technological Research Council of Turkey (TUBITAK) under Grant Number 124F213. The authors thank TUBITAK for their support. This work was also supported by the Scientific and Technological Research Council of Turkey (TUBITAK) under the 2211/C National Ph.D. Scholarship Program in Priority Fields in Science and Technology.

**Ethics approval and consent to participate**

The datasets used in this study are publicly available from open-access sources. All datasets are fully de-identified and contain no personally identifiable information. Therefore, ethics approval and consent to participate were not required for this study.

**Consent for publication**

Not applicable.

**Availability of data and materials**

The original dataset used in this study was obtained from the OASIS-1 (Open Access Series of Imaging Studies) project.

> Marcus DS, Wang TH, Parker J, Csernansky JG, Morris JC, Buckner RL (2007) Open Access Series of Imaging Studies (OASIS): Cross-sectional MRI Data in Young, Middle Aged, Nondemented, and Demented Older Adults. Journal of Cognitive Neuroscience 19:1498–1507. https://doi.org/10.1162/jocn.2007.19.9.1498.

The dataset was processed and converted from .nii to .jpg format and obtained from the version available on the Kaggle platform:

> AITHAL N (2025) OASIS Brain MRI Dataset (JPEG Format) for Alzheimer’s Disease Detection. https://www.kaggle.com/datasets/ninadaithal/imagesoasis/data (accessed 21 September 2026).

The external validation dataset used in this study was the ADNI: Alzheimer's MRI Classification Dataset.

> Alzheimer’s Association, Stages of Alzheimer's, Alzheimer’s Association Website (2025), https://www.alz.org/alzheimers-dementia/stages (accessed 21 September 2026).

**Code availability**

The code supporting the findings of this study is not publicly available. However, it can be obtained from the corresponding author upon reasonable request.

**Acknowledgments**

This research used resources of the National Energy Research Scientific Computing Center, a DOE Office of Science User Facility supported by the Office of Science of the U.S. Department of Energy under Contract No. DE-AC02-05CH11231 using NERSC award NERSC DDR-ERCAP0033396 and DDR-ERCAP0038319.

**Competing interests**

The authors declare no conflicts of interest.